\documentclass[conference]{IEEEtran}

\usepackage{times}

\usepackage[numbers]{natbib}
\usepackage{multicol}
\usepackage[bookmarks=true]{hyperref}
\usepackage{amsmath,amssymb,amsfonts}
\usepackage{amsthm}
\usepackage{algpseudocode}
\usepackage{graphicx}
\usepackage{float}
\usepackage{subcaption}
\usepackage{textcomp}
\usepackage{xcolor}
\usepackage{balance}
\usepackage{lipsum}
\usepackage{cuted}

\usepackage{enumitem}
\usepackage[english]{babel}
\usepackage{algorithm}
\usepackage{algpseudocode}
\newtheorem{theorem}{Theorem}

\newtheorem{remark}{Remark}

\usepackage{booktabs}  % For nicer horizontal rules
\usepackage{array}     % For better column formatting
\usepackage{xcolor} % For coloring rows (optional)

\newtheorem{definition}{Definition}
\DeclareMathOperator*{\argmax}{arg\,max}

\begin{document}

% paper title
\title{Behavior Synthesis via Contact-Aware Fisher Information Maximization}
% Behavior Synthesis via Contact-Aware fisher Information Maximization
\author{
  \IEEEauthorblockN{Hrishikesh Sathyanarayan and Ian Abraham}
  \IEEEauthorblockA{Yale University, New Haven, CT, USA \\
    Email: \{hrishi.sathyanarayan, ian.abraham\}@yale.edu}
    \vspace{-10em}
}

\maketitle

\begin{strip}
  \centering  \includegraphics[width=\textwidth]{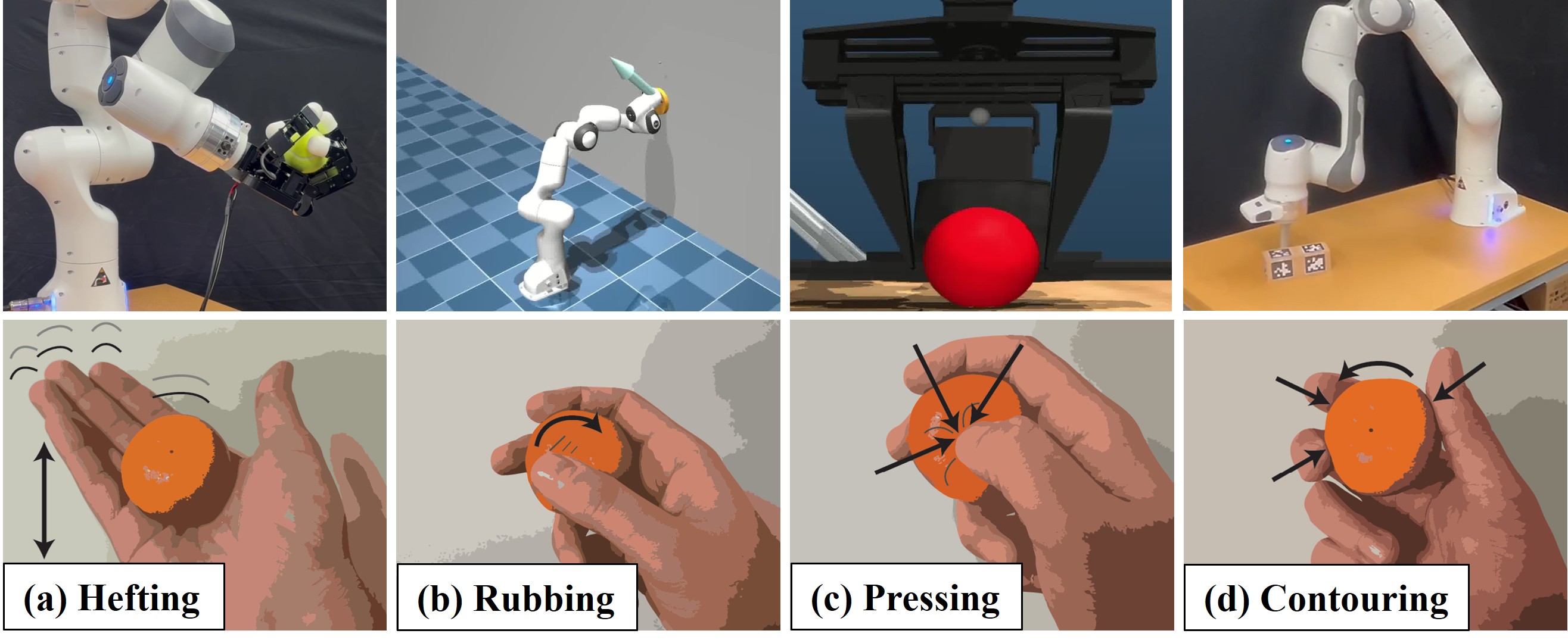}
  \captionof{figure}{\textbf{Emergent Contact-Based Learning Behaviors. } Here, we show emergent tactile behaviors resulting from the proposed contact-aware Fisher information maximization method that results in human-like tactile behaviors for learning (a) mass and weight, (b) friction and textures, (c) stiffness, and (d) shape \cite{LEDERMAN199329}. Additional media and code is provided in \url{https://github.com/ialab-yale/contact_aware_active_learning} .}
  \label{fig:abstract_figure}
\end{strip}

\begin{abstract}
    Contact dynamics hold immense amounts of information that can improve a robot's ability to characterize and learn about objects in their environment through interactions.
    However, collecting information-rich contact data is challenging due to its inherent sparsity and non-smooth nature, requiring an active approach to maximize the utility of contacts for learning.
    In this work, we investigate an optimal experimental design approach to synthesize robot behaviors that produce contact-rich data for learning. 
    % In this paper, we explore emergent learning behaviors resulting from actively planning and exciting contact modes that improve robot learning. 
    % ^ IA : This is really good, just wanted to parphrase a bit and give some context
    % The learned behaviors are synthesized from an optimal experimental design framework derived from information-theoretic methods that identifies and searches for information-rich contact modes through the use of contact-implicit optimization.
    Our approach derives a contact-aware Fisher information measure that characterizes information-rich contact behaviors that improve parameter learning. 
    We observe emergent robot behaviors that are able to excite contact interactions that efficiently learns object parameters across a range of parameter learning examples. 
    % robot exhibits from reasoning about unknown physics parameters that are learned from active interactions with the environment.
    % We show that by planning contacts in intentional and meaningful ways, the agent exhibits naturally emergent behaviors that gather diverse and information-rich contact datasets that efficiently estimate parameters of interest. 
    % We find significant improvement in parameter learning across a range of examples pertaining to learning object parameters. 
    Last, we demonstrate the utility of contact-awareness for learning parameters through contact-seeking behaviors on several robotic experiments.
    % Experiments were conducted in a range of manipulation-based settings and were shown to produce emergent learning behaviors that learn parameters more efficiently than random sampling baselines. 
\end{abstract}

\IEEEpeerreviewmaketitle

\section{Introduction}
    Contact dynamics are commonly used in robotics to manipulate the robot itself, e.g., through locomotion, or manipulate
    objects in its environment. 
    However, the utility of contacts goes beyond just manipulation, and instead, contact can be seen as a medium to transmit information that can help a robot learn about its environment.
    In fact, prior work has demonstrated the information-richness of contact as a means to improve parameter estimation problems \cite{nima,MI_review,Ogawa_humanoid}. 
    The underlying challenge is enabling robot behaviors that can actively acquire contact data for learning.     
    % IA -- ^ continue from this point
    
    % Robots often require making and breaking contact with the environment, especially with locomotion and dexterous manipulation. 
    % Such interactions ultimately influence the motion behaviors the robot exhibits during operation.
    % It is challenging, however, to plan and reason over desired contact modes due to the non-smooth and non-differentiable landscapes that cannot be simplified through linearization. 
    % Although explicitly defining a robot's contact sequence avoids this problem, it still has limitations due to overconstraining robot motion to abide to a strict contact schedule, as opposed to exploring for more diverse interactions.
    % However, through the use of contact-implicit optimization, we can plan over the brittle landscapes of contact and produce the desired contact modes for achieving tasks that require such interactions.

    Most prior work that leverages contact for learning acquire contact data as a by-product of implicit interactions \cite{nima,MI_review}. 
    However, information ``rich'' contact interactions are scarcely found through unintentional interactions, necessitating the need for robots to \emph{actively} explore contacts.
    Past work has shown the advantages of active exploration as a means to improve data collection for learning \cite{wilson_sac,631234}, but more work is needed to improve the quality of contact data collected in exploration. 
    In particular, some physical parameters of objects, e.g., shape and inertial parameters, can only be detected by continuously breaking and making contact \cite{directposa,ContactNets,yang2023impactinvariant,parmar2021fundamental,pmlr-v168-jin22a,10.1109/ICRA46639.2022.9811957}.
    \emph{Thus, the goal of this work is to develop the theoretical foundations that optimize for robot behaviors that produce contact-rich data for learning.} 

    To achieve this goal, we first formulate the maximum likelihood formulation for parameter learning subject to contact-based physics constraints. 
    In this formulation, we connect the robot's dynamics, and contact constraints, with the generation of sensor data for learning. 
    We employ techniques from optimal experimental design and derive a utility measure based on the Fisher information \cite{fedorov2010optimal,atanasov2013information} that explicitly takes into account information-rich sensor data as a function of contact. 
    % \begin{figure}[H]
    %     \centering
    %         \includegraphics[width=0.45\textwidth]{figs/abstract_figs/ABS-FIG-final_fig.png}
    %     \caption{\textbf{Example Contact-Aware Experimental Design for Parameter Learning Scenarios. }Top figure is the experimental setup for a `hefting' experiment, where the system exploits its contact interactions to estimate the mass of a block. Bottom figure demonstrates a bimanual Aloha 2 system estimating material parameters of a disk through contact exploration via contact-aware Fisher information maximization.}
    %     \label{fig:abstract_figure}
    % \end{figure}
    \noindent By maximizing this contact-aware Fisher information measure, we observe emergent robot behaviors that reasons about contact interactions that significantly improves parameter learning. 
    
    % \begin{figure}[t!]
    %     \centering
    %     \includegraphics[width=\linewidth]{figs/abstract_figs/ABS-FIG-final_fig.png}
    %     \caption{\textbf{Emergent Contact-Based Learning Behaviors. } Here, we show emergent behaviors resulting from the proposed contact-aware Fisher information maximization that are similar to human contact exploration \cite{LEDERMAN199329}. 
    %     Resulting behaviors facilitate learning (a) mass and weight, (b) friction and textures, (c) stiffness, and (d) shape. 
    %     % The top row shows the emergent learning behaviors when reasoning about different physical parameters like (a) mass, (b) friction coefficient, (c) material properties, and (d) material properties and friction coefficient. 
    %     }
    %     \label{fig:abstract_figure}
    %     \vspace{-2em}
    % \end{figure}

    We focus on a class of example problems where the robot is tasked to interact with and learn object properties through interaction (see Fig.~\ref{fig:abstract_figure}). 
    These types of problems are interesting because many physical parameters can only be estimated from physical interaction \cite{nima, MI_review}. 
    In this work, we show that a contact-aware optimal experimental design approach can produce \emph{meaningful} contact interactions for robots to obtain information-rich data that facilitates improvement in parameter learning. 
    In summary, the contributions of this paper are the following:

    \begin{enumerate}
        \item Formulation of a contact-aware Fisher information measure to identify information-rich contacts.
        \item Development of an optimal control approach that plans contact sequences that improves parameter learning.
        \item Demonstration of our approach generating emergent contact behaviors on a variety of robot parameter learning scenarios.
    \end{enumerate}

    This paper is structured in the following manner: Section \ref{sec:related_work} overviews prior related work. Section \ref{sec:prelims} overviews preliminary material on Parameter Estimation, Fisher information, and Experimental Design. Section \ref{sec:methods} details the methodologies used to construct our approach. Section \ref{sec:results_and_discussion} discusses our results and insights from using our approach. Section \ref{sec:lims} overviews limitations and future works. We conclude the paper in Section \ref{sec:conclusion}.

\section{Related Work}
\label{sec:related_work}

    \subsection{Parameter Estimation}
        
        Parameter estimation has been widely explored where unknown physical parameters about the robot need to be estimated more precisely through measurements \cite{MI_review,194738,4048619,10.5555/293154}. 
        % or the environment are unknown a priori, and the agent must make meaningful interactions with the environment in order to properly identify the parameters of interest.
        In general, parameter estimation involves utilizing sensor measurements from a system in order to fit unknown parameters of interests to the sensor readings based on a model of choice that captures the dynamical behaviors of the system. 
        In the context of learning parameters from contact, prior work remarks on the observability and identification of parameters \cite{nima}, stating that not all interactions a system makes can be useful for estimating parameters. For example, if a system makes no effort to interact with the environment, and vice versa, parameters such as mass and inertia are indistinguishable due to the homogeneity of the equations of motion, and can only retrieve a ratio of the parameters.

        Parameter estimation algorithms generally consist of fitting parameters comprised in a model structure with sensor readings from the robot. Depending on the type of sensor used, certain classes of physical parameters can be reasonably identifiable \cite{BELLMAN1970329,LINDER20146454}.  For instance, estimating inertial parameters becomes intractable if the only available sensor for parameter estimation is an accelerometer, and perceived contact forces have no reference value to account for inertia. Therefore, choosing the \emph{type}  of sensors available for parameter estimation is crucial before deducing what unknown parameters should be estimated.

        Common methods for parameter estimation are Bayesian-inference based approaches \cite{PEPI2020103025,rothfuss2024bridgingsimtorealgapbayesian} and least squares estimation \cite{nima}. Bayesian approaches for parameter estimation bake parameter uncertainty into the problem and can generalize well for complex, nonlinear problems, at the cost of computational efficiency for large data. Least squares estimation works well for simple, linear regression problems, but struggles to generalize to complex, nonlinear problems. 
        Other methods are presented in \cite{sundaralingam2021inhandobjectdynamicsinferenceusing}, where parameter estimation is solved via a factor graph-based approach. 
        However, factor graph-based methods are computationally heavy to compute, and requires careful engineering to construct.
        While there are many approaches to parameter estimation, this paper formulates the parameter estimation as a maximum likelihood problem \cite{ge2023maximumlikelihoodestimationneed}, which allows us to consider the dependence of data in parameter estimation.
        Specifically, the maximum likelihood formulation is amenable to techniques in optimal experimental design that provides meaningful requirements for measurement data. 
        % capable of estimating parameters for nonlinear systems given measurement data with high noise. Such an approach is also useful when structuring and optimizing over Fisher information, discussed later.
        
    \subsection{Optimal Experimental Design and Fisher Information}

        The study of optimal experimental design \cite{AF_Emery_1998,fedorov2010optimal} provides an insight towards how to best sample data for estimating parameters in a maximum likelihood problem. Experimental design involves optimizing over an information metric in order to solve for expected measurements in an experiment that yield the most \emph{information-rich} data about unknown parameters of interest. Fisher information is often used in experimental design, as it directly captures how sensitive the sensor measurements are to slight perturbations of the parameter estimates.
        % IA Add in intuition for these methods and why they work
        By maximizing Fisher information \cite{atanasov2013information,AF_Emery_1998}, we take an outer product of the gradients over the maximum likelihood reward function with respect to parameters of interest and solve for regions in this new reward landscape where the function is the \emph{most sensitive} to slight perturbations of the parameters. This optimization problem is known as experimental learning. The Fisher information metric is often used as it captures information-richness in datasets and is directly related to the variance of parameter estimates of interest \cite{Rao_1947}. By optimizing directly over the Fisher information landscape, we yield experiments that lead to desirable, information-rich data \cite{wilson_fishermax,wilson_sac}.
    
        In the context of robotics, experimental learning provides us a means to acquire information-rich data that describes unknown parameters of interest accurately \cite{wilson_fishermax,wilson_sac}. The learning algorithm will output a control sequence that is applied to the robot which generates a motion that collects information-rich data through equipped sensors.

    \subsection{Contact Implicit Trajectory Optimization}

        This paper takes a conventional experimental design problem and primes the Fisher information with contact-awareness, allowing the robot to optimize for \emph{information-rich} contact interactions to improve parameter learning. In doing so, we pose an experimental design problem where contact is an implicit variable for optimization as a function of robot trajectory. Contact-implicit trajectory optimization \cite{Manchester-2017-122109,10.1145/2185520.2185539,6386025} synthesizes trajectories wherein contact forces, robot state, and control inputs are posed as decision variables in the optimization problem. 
        They are generally solved by optimizing for robot trajectory and contacts over an objective, with constraints being robot dynamics and a contact model of choice \cite{directposa,nima,ContactNets,6386025}. 
        Essentially, these methods can generate motions without explicitly specifying contact mode sequences.

        Such formulations of optimization problems have been shown to have impressive results, shown in \cite{10.1145/2185520.2185539,6386025,directposa}. 
        These problems are formulated by a choice of contact model and optimization methods. 
        Particular to the algorithm presented in \cite{directposa}, the resulting motion is generated through a time stepping method with linear complementarity constraints posed by \cite{Stewart2000AnIT,Anitescu} in order to resolve rigid-body contact modes implicitly. 
        % Other works seek to optimize contact modes with smoother models presented in \cite{6386025} and solve using indirect Differential Dynamic Program-based (DDP) methods \cite{doi:10.1080/00207176608921369,7845678}.
        In this work, we address a parameter learning problem formulated as an \emph{contact-implicit optimization}, using a contact model of choice, where the parameters are embedded within the contact model. We aim to refine parameter estimates by  optimizing contact modes such that they align closely with given contact data.

    \subsection{Belief Space Planning and Control}

        Belief space planning is commonly used in robotics to plan under parameter uncertainty by considering the robot's belief about the parameter, and gathering information to converge on parameter uncertainty. 
        Consequently, the robot infers unknown properties of objects around it and plans for interactions accordingly. 
        Early works in \cite{Platt2010BeliefSP,Lauri_2016,KAELBLING199899} present belief space control using a Partially Observable Markov Decision Process (POMDP), where the robot explores for relavant and rich information about parameters which is used to solve a maximum likelihood problem.
        
        Recent approaches \cite{tremblay2023learningactivetactileperception,memmel2024asidactiveexplorationidentification} leverage the belief space formulation for improving contact-based robot tasks.
        Our approach is similar in that we leverage the parameter uncertainty based on the sensitivity of the measurement model to improve robot task performance. 
        The main difference is 1) the explicit consideration of contact mechanics in computing the Fisher information; and 2) that we can solve the contact-aware Fisher information maximization problem via sample-based control for real-time performance. 
        The advantage is that we can generalize across a larger range of contact-based robotics tasks, and that can readily solve the information maximization problem under real-time control setting without needing to resort to computationally expensive reinforcement learning methods.
        % using traditionally Reinforcement Learning (RL) methods under practical robotic settings, where the robot plans trajectories with expected rich-information in simulation and executes actions on the real robot, thus improving downstream learning tasks. However, such approaches approximate gradients of the Fisher information using a finite difference approximation that smooths out the gradients of the Fisher information. We propose a sampling-based control method that structures the measurement model with an explicit consideration of contact, such that the robot plans to learn parameters by reasoning over its contact interactions more carefully.

\section{Preliminaries}
\label{sec:prelims}

    \subsection{Parameter Learning via Maximum a-posteriori Estimate}

        In this work, we cast parameter learning as a maximum a-posteriori estimation over parameters $\theta \in \Theta$. 
        Let $p(\theta) : \Theta \to \mathbb{R}$ be the prior distribution over $\theta$ and $\mathcal{D} = \{(x_i, y_i) \}_{i=1}^N$ be a collection of $N$ input-output sensor data where $x_i \in \mathcal{X}$ is an input, and $y_i \in \mathcal{Y}$ is noisy sensor data.  
        Given a likelihood model\footnote{In our work, we call this the sensor model which provides the likelihood that the collected data $\mathcal{D}$ is explained by parameter $\theta$.} $p(\mathcal{D} | \theta)$, the posterior belief $p(\theta | \mathcal{D})$ is given by the Bayesian update
        \begin{equation} \label{eq:marginilization_theta}
            p(\theta | \mathcal{D}) = \frac{p(\mathcal{D} | \theta) p(\theta)}{\int_\Theta p(\mathcal{D} | v)p(v) dv}.
        \end{equation}
        It is often difficult or intractable to directly solve for the posterior distribution given data. 
        Instead, it is easier to just optimize for the mode $\hat{\theta}$ of the posterior distribution. 
            
        % In this work we consider the following linear regression model for a single observation: $y = \phi(x_i)^T \theta + b$, where $y \in \mathbb{R}$, $\phi : \mathbb{R} \rightarrow \mathbb{R}^{p}$, $\theta \in \mathbb{R}^{p}$, $b \in \mathbb{R}$. Here, $y$ is an observed output from $N$ observations, where total measurements $Y = [y_1 \dots y_N] \in \mathbb{R}^N$, $\phi(x^k)$ is our feature that maps an observation to $p$ parameters, $\theta$ comprises $p$ unknown parameters, and $b$ is our iid Gaussian noise. We represent a stack of features as our design matrix, $\mathbf{\Phi}(\mathcal{D}) = [\phi(x_1) \dots \phi(x_N)] \in \mathbb{R}^{p \times N}$, where $\mathcal{D}$ is our dataset measured from an experiment of $N$ observations.

        \begin{definition}\label{def:mle}
            \textbf{Maximum a-posteriori Estimation.} Given a prior $p(\theta) : \Theta \to \mathbb{R}$ over parameters $\theta \in \Theta$, a data set of $N$ independent and identically distributed (i.i.d.) measurements $\mathcal{D} = \{(x_i, y_i) \}_{i=1}^N$, and a likelihood model $p(\mathcal{D} | \theta)$, the point estimate of the posterior distribution is given as the maximum a-posteriori (MAP) estimate $\hat{\theta}$ which is the solution to the optimization 
            \begin{equation}\label{eq:regularized_maxlikelihood}
                \hat{\theta} = \argmax_{\theta} p(\mathcal{D} | \theta) p(\theta).
            \end{equation}
        %     \begin{equation}\label{eq:regularized_maxlikelihood}
        %         \theta^* = \argmax_{\theta} \textrm{log} \prod_{i=1}^{T} p(y_i\ |\ x_i, \theta) p(\theta)
        %     \end{equation}
        \end{definition}
          In practice it is common to optimize for the log a-posteriori $\hat{\theta}=\argmax_{\theta} \log \left( p(\mathcal{D} | \theta) p(\theta) \right)$ which often has better numerical conditioning. 
        % The condition $\mathbb{E}\left[p(\mathcal{D}|\theta)\right] = \hat{\theta}$, where $\hat{\theta}$ is an estimate of $\theta$, is assumed to be satisfied in order for the likelihood model to be an unbiased estimator of $\theta$.
        % Note that parameters $\theta$ is not known, and the measurements we obtain rely heavily on such model parameters. As indicated in Eq. \ref{eq:regularized_maxlikelihood}, we numerically stabilize the parameter estimation by marginalizing over parameters $\theta$ that maximizes the likelihood of output $y_i$ solely given observation $x_i$, given a gaussian prior of $\theta$, by using the following Bayesian update, 

        % \begin{equation} \label{eq:marginilization_theta}
        %     p(\theta| y_i, x_i) = \frac{p(y_i\ |\ x_i, \theta) p(\theta)}{\int{p(y_i\ |\ x_i, \theta) p(\theta)}d\theta}
        % \end{equation}
        
        % In maximum likelihood optimization formulation in Eq. \ref{eq:regularized_maxlikelihood}, we can see that the denominator term in Eq. \ref{eq:marginilization_theta} is integrated out to be a constant. Using a prior of $\theta$ allows for regularization of the parameter estimates based on a prior average belief of the estimates, allowing for numerically stable gradients when solving for parameter estimates.

        \begin{remark}
            Not including the prior $p(\theta)$ in the maximization in~\eqref{eq:regularized_maxlikelihood} yields the simpler, but effective maximum likelihood problem. While the formulation remains useful for accounting for collected data, it is unable to account for prior information.
        \end{remark}

        In general, solving the maximum a-posteriori estimation assumes data has been provided. 
        % is used to fit the best parameter $\theta$ to the data, but itself does not provide the data (this is usually given). 
        We are interested in producing a method that provides the best data $\mathcal{D}$ that improves the conditioning of~\eqref{eq:regularized_maxlikelihood}.
        
    \subsection{Fisher Information}

        One way to establish better conditioning on maximum a-posteriori problems is to analyze the statistical manifold of the problem. 
        In particular, we are interested in analyzing the sensitivity of the posterior distribution with respect to the parameter.
        The insight is that certain measurements yield more \emph{information} about parameters, i.e., reduction in uncertainty, when the gradient of the likelihood model is largest. 
        A common measure of information with respect to parameter sensitivity is the \emph{Fisher information matrix}. 
        
        % In general, Fisher information is a Riemannian metric over statistical manifolds that quantifies information-richness over the observations in an experiment. This paper considers Fisher information as a metric that models the sensitivity of our parameters $\theta$ with respect to the measurement outputs $y_i$. The notion of parameter sensitivity is crucial because we can optimize over $\theta$-relevant information spaces where our measurement model is changing the \emph{fastest} over small perturbations of parameter values $\theta$ (see Section \ref{subsec:exp_design}).
        \begin{definition}\label{def:empricial_fim}
            \textbf{Fisher Information Matrix.} Let $\mathcal{L}(\mathcal{D} | \theta)$ be the log likelihood of an unbiased estimator that is continuous and differentiable with respect to elements in $\mathcal{D}$ and $\theta \in \Theta \subseteq \mathbb{R}^d$.  
            Then, the Fisher Information Matrix $\mathcal{F}(\mathcal{D} | \theta) \in \mathbb{R}^{d \times d}$  of $\theta$ is defined by 
            % assume it is  and $\mathcal{D}$ is a random variable. Given a log product of a measurement model $\log{\prod_{i=1}^N p(y_i\ | x_i, \theta)}$ for all observations $x_i$ at instances $i \in [0,N]$, the Fisher information is modeled as the Hessian of the log product measurement model or the outer product of its score,
            \begin{equation}\label{eq:general_fisherinf}
            % \begin{split}
               \mathcal{F}(\mathcal{D} | \theta) =\mathbb{E}\left[\frac{\partial}{\partial \theta} \mathcal{L}(\mathcal{D} | \theta) \frac{\partial}{\partial \theta} \mathcal{L}(\mathcal{D} | \theta)^\top\right]
               % &= \mathbb{E}\left(\nabla_\theta \log{\prod_{i=1}^N p(y_i\ | x_i, \theta)} 
               %  \log{\prod_{i=1}^N \nabla_\theta p(y_i\ | x_i, \theta)^T}\right)
            % \end{split}
            \end{equation}
            and 
            \begin{equation}
                \mathcal{F}(\mathcal{D} | \theta) = - \mathbb{E} \left[ \nabla_\theta^2 \mathcal{L}(\mathcal{D} | \theta) \right]
            \end{equation}
            if $\mathcal{L}\in\mathcal{C}^2$. Here, the expectation is with respect to the random variable $\mathcal{D}$ and $\nabla_\theta^2$ is the Hessian~\cite{c9b3ce2a-e311-3788-ad85-e058d2df325b}.
            % If $\mathcal{L}$ is twice differentiable, it is common to also define the Fisher information matrix as 
            % \begin{equation}
            %     \mathcal{F}(\mathcal{D} | \theta) = - \mathbb{E} \left[ \nabla_\theta^2 \mathcal{L}(\mathcal{D} | \theta) \right]
            % \end{equation}
            % where $\nabla_\theta^2$ is the Hessian which is equivalent to the prior expression (see \cite{c9b3ce2a-e311-3788-ad85-e058d2df325b}).
        \end{definition}
        % Note that the dataset $\mathcal{D} = \{ x_i \}_{i=1}^N$ comprises of all observations in an experimental run. Intuitively, Fisher information determine how quickly the log-likelihood model is changing with respect to $\theta$, allowing us to quantify the sensitivity of our likelihood model with respect to the parameters. 
        Interestingly, the Fisher information matrix also has a unique interpretation as a lower-bound on the uncertainty of the posterior in maximum likelihood problem. 
        \begin{definition}\label{def:CRLB}
            \textbf{Cram\'er-Rao Lower Bound \cite{Rao_1947}. }
            Let $\mathcal{D}$ be a random variable drawn from a distribution $p(\mathcal{D} | \theta)$, and $\hat{\theta}$ be the unbiased estimator for $\theta$, then 
            % Fisher information is also closely related to the variance of the parameter estimates $\theta$ by the Cramer-Rao lower bound \cite{Rao_1947}, written as,
            \begin{equation}\label{eq:cramer_rao}
                \mathcal{F}(\mathcal{D} | \theta) \geq \text{cov}(\hat{\theta})^{-1}.
            \end{equation} 
        
        \end{definition}
        For the interested reader, the proof of this lower bound can be found in \cite{Rao_1947}. 
        The bound shows that maximizing the Fisher information matrix directly impacts the variance of the parameter estimate which we use to establish an optimization problem over \emph{planned datasets}.
        % Consequently, by increasing the determinant (or trace) of the Fisher information, parameter uncertainty will collapse accordingly. 
        
    \subsection{Optimal Experimental Design}\label{subsec:exp_design}

        We can maximize the Fisher information in order to choose measurements $\mathcal{D} = \{ (x_i, y_i)\}_{i=1}^N$ that improves the precision of parameter estimation. 
        The idea is to establish an optimization on the input design of an experiment, i.e., the independent variable $x_i$, such that the dependent variable $y_i$ that has not been collected has more utility. 
        To establish this optimization, we first assume maximum likelihood and normally distributed measurements\cite{zheng2021beliefspaceplanningcovariance} where $\mathcal{L}(\mathcal{D}|\theta) = \mathcal{L}(X | \theta)$ and $X = \{x_i\}_{i=1}^N$.
        Under a finite sampling of independent variables $x_i$ we obtain the empirical Fisher information  
        \begin{equation}
            \begin{split}
                \mathcal{F}(\mathcal{D} | \theta) &\approx - \frac{1}{N}\sum_{i=1}^N \frac{\partial^2 \mathcal{L}(X_i|\theta) }{\partial \theta^2} \\
                & = \frac{1}{N}\sum_{i=1}^N \frac{\partial \mathcal{L}(X_i | \theta)}{\partial \theta}
                \frac{\partial \mathcal{L}(X_i | \theta)}{\partial \theta}^\top.
            \end{split}
        \end{equation}
        
        Since $X$ is now an independent control variable, the following \emph{optimal design} is constructed
        \begin{equation}\label{eq:exp_design_formulation}
            \max_{X} \psi\left(\mathcal{F}( X| \theta)\right)
        \end{equation}
        where the Fisher information matrix is overloaded with argument $X$ \cite{Rao_1947,fedorov2010optimal}, and $\psi(\cdot)$ is a matrix reduction operator over the Fisher information that returns a scalar measure of the matrix, e.g., the determinant (D-optimality), or the trace (T-optimality). 
        Note that the Fisher information is symmetric and positive semi-definite \cite{https://doi.org/10.1111/j.2517-6161.1959.tb00338.x,ChirikjianGregoryS2009SMIT}.
        The inducing operator $\psi$ is then a metric on the information landscape that alter the experiment design, e.g., D-optimality will maximize the total volume of the Fisher information matrix, whereas T-optimality maximizes the largest enclosing sphere.  
        % computationally efficient and still leads to a collapse of parameter uncertainty. 
        % The goal of this paper is to leverage this formulation to facilitate contact reasoning for parameter learning. 
        Our goal is to formulate a \emph{contact-aware} formulation of experimental design to facilitate with contact-based robot tasks that seeks information-rich contact for learning.

    \begin{figure*}[t!]
            \centering
            \includegraphics[width=\textwidth]{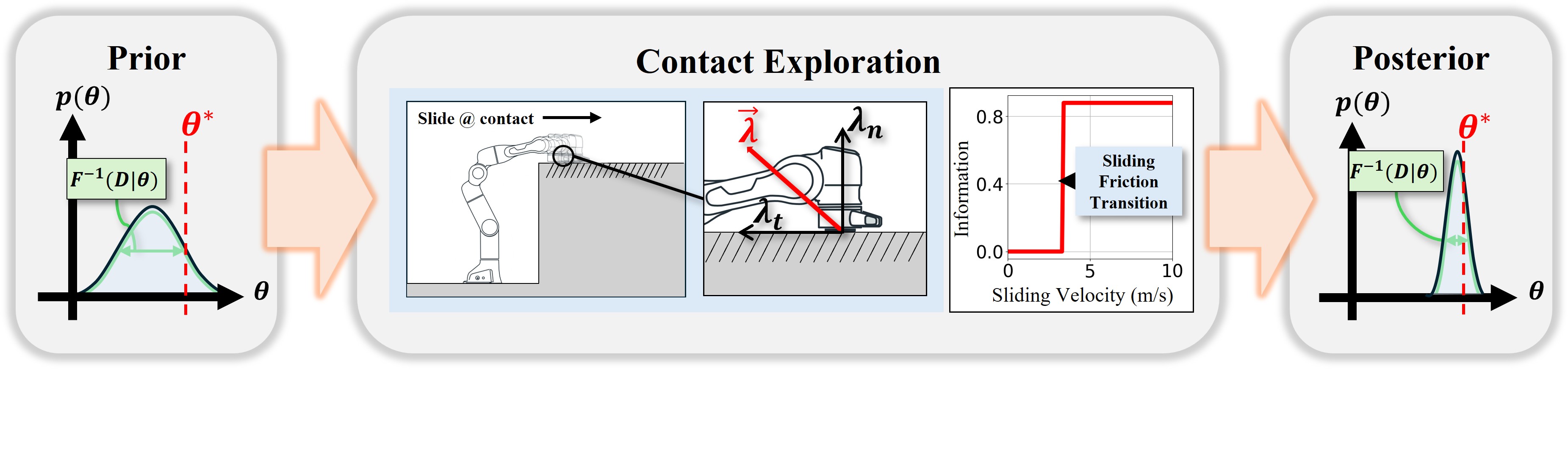}
            \caption{\textbf{Collapsing Parameter Uncertainty Through Contact. } In this example, the robot seeks to estimate the friction coefficient, $\mu$, of a nearby table. Given an initial prior of the parameter estimate, the robot seeks to update its belief of the friction coefficient by interacting with the environment through contact. In doing so, the robot collects contact sensor measurements that hold information, $\mathcal{F}$ (green curve), about the parameter, which is then used to update the parameter estimate and variance. We can see that in order to reduce parameter variance and converge to a more precise parameter estimate, the robot exhibits a \emph{sliding} behavior that excites terms in the Fisher information, which is a \emph{lower bound} on parameter uncertainty \cite{Rao_1947}.}
            \label{fig:intuition}
    \end{figure*}  

\section{Contact-Aware Experimental Design}
\label{sec:methods}

    To enable robots to actively seek out contact, we need to derive a notion of ``information'' related to parameter estimation subject to contact.
    Thus, we first need to establish the explicit dependence of contact in parameter estimation. 
    
    \subsection{Contact-Aware Max. A-Posteriori Estimation (CA-MAP)} 
    \label{subsec:methods_problem_formulation}

        Let us define a robot with the state variable $x_t \in \mathcal{X}$ with control input $u_t \in \mathcal{U}$ at some discrete time $t \in \mathbb{Z}^+$. 
        Additionally, let us define the dynamics constraints of the robot as $x_{t+1} = f_\theta(x_t, u_t, \lambda_t)$ and the sensor measurement model $y_t=g_\theta(x_t)$ parameterized by $\theta \in \Theta$ where $y_t \in \mathcal{Y}$ is a sensor measurement, and $\lambda_t \in \mathcal{C}_\theta(x_t)$ is the contact constraint. Here, $\mathcal{C}_\theta(x)$ denotes the contact conditions as a function of state, e.g., friction, non-penetration, and complementarity.
        % Next, let us consider the dynamic constraints of the robot as $x_{t+1} = f(x_t, u_t, \lambda_t)$ and the sensor measurement model $y_t=g_\theta(x_t)$ parameterized by $\theta$ where $y_t \in \mathcal{Y}$ is a sensor measurement.
        We define an ``experiment'' as a set of $T$ measurements $\mathcal{D} = \{(y_t, u_t) \forall t \in [0,T-1]\}$ where the goal is to maximize the a-posteriori estimate $\theta$ given a prior $p(\theta)$ and experiment $\mathcal{D}$ subject to the dynamic, contact, and sensor model constraints.
        \begin{definition} \label{def:cmap} \textbf{Contact-Aware MAP (CA-MAP)}
            Let $\tau = \{ (x_t, \lambda_t) \forall t \in [0,T-1] \}$ be a trajectory of states $x_t$ and contact forces $\lambda_t$ for some time horizon $T\in\mathbb{Z}^+$. Given the likelihood model $p(\mathcal{D}| \tau, \theta)$ (i.e., a sensor model) that is normally distributed with mean $g_\theta(x)$ and variance $\Sigma$, the contact-aware maximum a-posteriori estimation (CA-MAP) problem is defined as
            \begin{equation}
                        \label{eq:contact_aware_max_like}
                \begin{aligned}
                    & \  
                    \hat{\theta} = \argmax_{\theta, \tau} \log p(\mathcal{D} | \tau, \theta) p(\theta)  \\ 
                    \text{s.t.} & 
                    \begin{cases}
                        x_0 \text{ given } & \text{ (initial cond.) } \\ 
                        x_{t+1} = f_\theta(x_t, u_t, \lambda_t) & \text{ (dynamics model) } \\ 
                        \lambda_t \in \mathcal{C}_\theta(x_t) & \text{ (contact constraints) }  \\ 
                        \theta \in \Theta & \text{ (feasible parameter) }
                    \end{cases} \\
                \end{aligned}
            \end{equation}
            where the subscript $\theta$ denotes the functional dependence of the model on the parameter.
        \end{definition}

        The intuition behind this problem formulation is that we are optimizing for the parameter estimation \emph{and} the trajectory $\tau$ that best explains the data $\mathcal{D}$.
        The role of the contact model for $\lambda$ is to explicitly consider contact constraints as part of the parameter estimation problem. 
        In many scenarios, contact forces are not explicitly measured, but do appear in the physical constraints of the system and are critical for parameter estimation. 
        Thus, incorporating the contact model enables us to explicitly optimize contact interactions for experimental design. 
        % the are the unknown physics parameters that we are solving for, while solving for states $x_t \in \mathcal{X}$ and contacts $\lambda_t \in \mathcal{C}(x_t, u_t, \theta)$ that fit the sensor readings $\{ \bar{y}_t \}_{t=0}^T$. 
        % Here, $x_0$ and $u_t \in \mathcal{U}$ are the initial state and control sequence applied to the robot, respectively. We measure $\{ \bar{y}_t \}_{t=0}^T$ from the robot after running open loop controls $u_t \in \mathcal{U}$ with initial condition $x_0$.
        % We constrain the problem to the robot dynamics model $f$ that outputs the next state $x_{t+1}$ given input state, controls, and contact forces as inputs.
        % We define $p(y | x, u, \lambda, \theta)$ as the measurement likelihood model that captures the uncertainty in the measurements, with the mean $g(x, u, \lambda)$ being the sensor model. 
        % This paper seeks to leverage the contact-aware maximum likelihood formulation in order to produce a control sequence $u_t \forall t \in [0,T]$ that generates sensor readings $\{ y_t \}_{t=0}^T$ that excites the most informative contacts that improves learning $\theta$.

        \subsection{CA-MAP Fisher Information Matrix}
        Given the contact-aware a-posteriori estimation problem, our goal is to derive an equivalent notion of the Fisher information that explicitly considers contact.
        To achieve this, we first define the Lagrangian of the contact-aware parameter estimation problem as 
        \begin{equation}
            \mathcal{L}(\mathcal{D} | \tau, \theta, \alpha) = \ell(\mathcal{D} | \tau, \theta) + \alpha d(\tau, \theta)
        \end{equation}
        where $\alpha$ is the Lagrange multiplier for each constraint $d(\tau, \theta)$ and  $\ell(\mathcal{D} | \tau, \theta) = \log p(\mathcal{D} | \tau, \theta) p(\theta)$. 
        \begin{theorem}\label{theorem:fisher_max_step}
            Let $\mathcal{L}(\mathcal{D} | \tau, \theta, \alpha)$ be the Lagrangian of the contact-aware estimation problem~Def.~\ref{def:cmap} where $\mathcal{L} \in \mathcal{C}^2$.
            The contact-aware Fisher information matrix $\mathcal{F}(\mathcal{D} | \tau, \theta, \alpha) \in \mathbb{R}^{d \times d}$ for $\theta \in \Theta \subseteq \mathbb{R}^d$ is given by 
            \begin{equation} \label{eq:contact_aware_fish}
                \mathcal{F}(\mathcal{D} | \tau, \theta, \alpha) = -\nabla_\theta^2 \mathcal{L}(\mathcal{D} | \tau, \theta, \alpha) 
            \end{equation}
            defined by the steepest ascent problem 
            \begin{equation}
            \label{eq:steepest_asc_problem}
                \delta \theta^\star = \argmax_{\delta \theta} \nabla_\theta \mathcal{L}(\mathcal{D} | \tau, \theta)^\top \delta\theta + \frac{1}{2} \delta \theta ^\top \nabla_\theta^2 \mathcal{L}(\mathcal{D} | \tau, \theta) \delta \theta
                % - \gamma\Vert \delta \theta\Vert^2_{\Sigma_\theta^{-1}}
            \end{equation}
            where $\mathcal{F}(\mathcal{D} | \tau, \theta) = -\nabla_\theta^2 \mathcal{L}(D | \tau, \theta)$ and the dependence of the Lagrange multipliers are dropped and assumed to be constants.  
            % with respect to unknown parameters $\theta \in \Theta$ produces data that yields the steepest possible descent in $\Theta$ space for parameter learning.
        \end{theorem}
        \begin{proof}
            See Appendix \ref{app:steepest_asc_proof} for the complete proof.
        \end{proof}
        
        Theorem~\ref{theorem:fisher_max_step} connects a variation of the Fisher information matrix to a Newton's method conditioning for the steepest descent problem.
        Note that in this case, we define the empirical Fisher information (where the summation over data is implicit in the augmented log-likelihood model). 
        Furthermore, consider that the Fisher information has as an argument the trajectory $\tau = \{ (x_t, \lambda_t)\}_{t=0}^T$, and the measured data $\mathcal{D} = \{(y_i, u_i)\}_{i=0}^T$. 
        We can use this fact to \emph{solve} for robot behaviors that maximize the Fisher information, ultimately improving the conditioning on the steepest descent for the CA-MAP problem in the \emph{data} space where contact can be explicitly sought to excite information about physical parameters.

    \section{Contact-Aware Optimal Experimental Design}

        We now derive the contact-aware optimal experimental design approach to synthesize robot behaviors for learning. 
        Our approach integrates the contact-aware Fisher information into a closed-loop optimal design procedure to facilitate robot parameter learning from information-rich contact data. 

        \subsection{Contact-Seeking Behavior Synthesis} 
            We first establish the optimization formulation that maximizes the contact-aware Fisher information matrix. 
            \begin{definition} \textbf{(Contact-Aware Optimal Experimental Design (CA-OED))}.
                Let $\psi : \mathbb{R}^{d\times d} \to \mathbb{R}$ define a continuous and differentiable metric and $\mathcal{J}(\mathcal{D},\tau)$ denote a cost term on the trajectory.
                Additionally, assume the contact-aware Fisher information matrix $\mathcal{F}(\mathcal{D} | \tau, \theta)$ is symmetric and positive semi-definite~\cite{soen2021variancefisherinformationdeep,pmlr-v70-sun17b}.
                Then, for a given estimate $\theta$, the contact-aware optimal experimental design problem is defined as 
                \begin{align} \label{eq:exp_learning_opt}
                    \hat{\mathcal{D}}, \hat{\tau} = \argmax_{\mathcal{D}, \tau} \psi\left(\mathcal{F}(\mathcal{D}| \tau, \theta)\right) -\mathcal{J}(\mathcal{D}, \tau)\\\text{ s.t. }
                    \begin{cases}
                        x_0 \text{     (given)}\\
                        x_{t+1} = f_\theta(x_t,u_t,\lambda_t), \\
                        \lambda_t \in \mathcal{C}_\theta(x_t)  \\
                        y_t = g_\theta(x_t)
                    \end{cases} \nonumber
                \end{align}
                where $\hat{\mathcal{D}} = \{(y_t, u_t)\}_{t=0}^T$ is the predicted dataset and $\hat{\tau} = \{(x_t, \lambda_t)\}_{t=0}^T$ is a reference trajectory.
            \end{definition}

            We can solve this optimization problem in~\eqref{eq:exp_learning_opt} using any number of optimization techniques. We opted to solve this problem using predictive sampling \cite{howell2022predictivesamplingrealtimebehaviour} where contact constraints are implicitly solved through the dynamics model \cite{directposa,nima}.
            In this formulation of optimal experimental design, it is important to recognize that the measurements $y_t$ that are optimized are only predictions. 
            The purpose of the prediction is to specify the Fisher information matrix which explicitly considers the parameter sensitivity to the measurement. 
            As such, we can use the sensor predictions as an interpretation of what ``information-rich'' contact behaviors produce. 

            In this work, we use the trace for $\psi$ as the metric on the Fisher information.        By maximizing the trace Fisher information matrix, we improve the condition number of the matrix used for Newton's step in the CA-MAP problem thus allowing empirically faster convergence and numerical stability.
            An example of this phenomena is illustrated in Fig.~\ref{fig:intuition}.
    
        \begin{figure}[h]
            \centering
            \includegraphics[width=\linewidth]{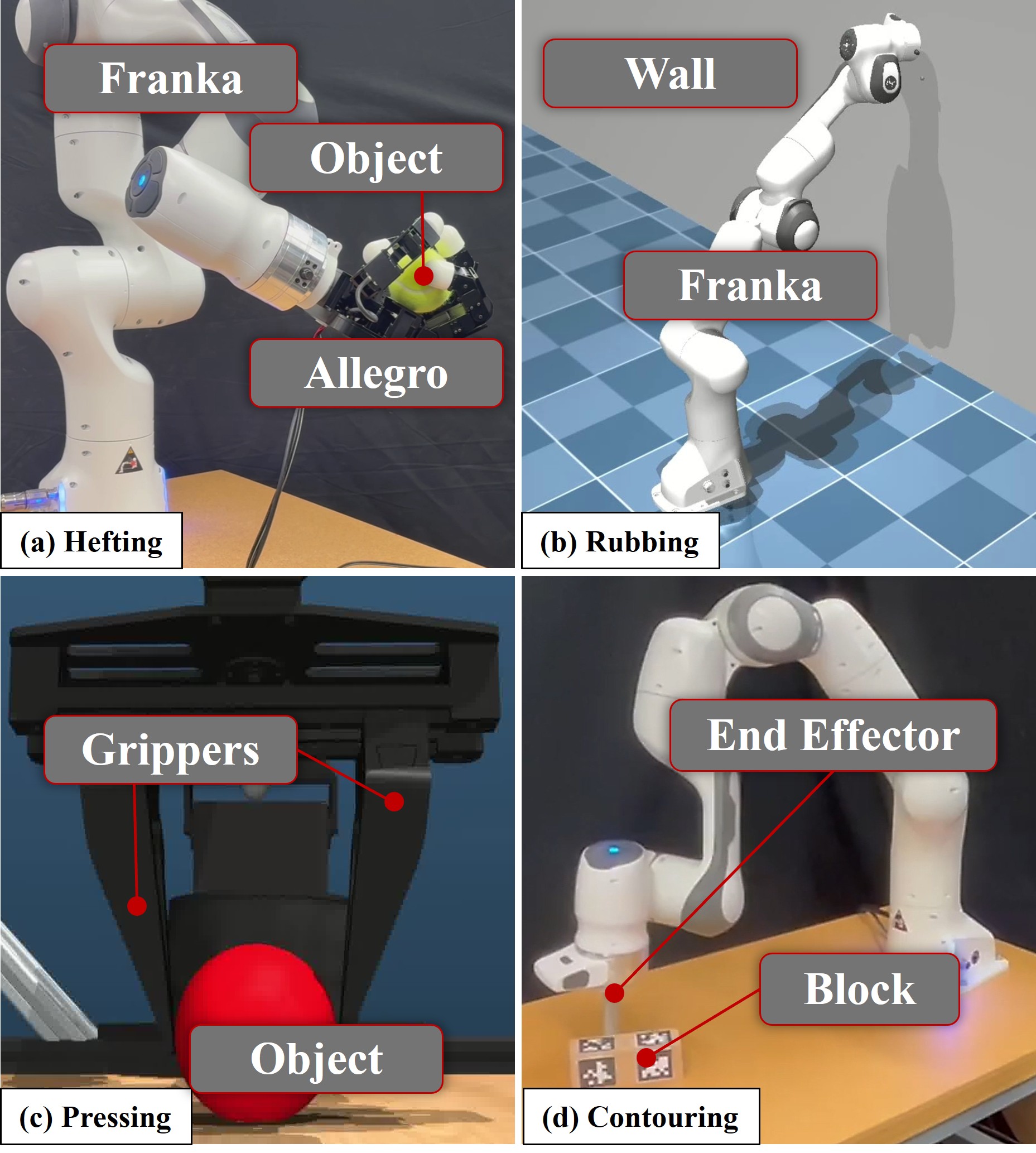}
            \caption{\textbf{Experimental Setup. } The experimental setup for (a) hefting experiments for mass estimation, (b) rubbing experiments for friction coefficient estimation, (c) pressing experiments for material stiffness and dampening estimation, and (d) contouring experiments for shape estimation.}
            \label{fig:exp_setup}
        \end{figure} 
    
        \subsection{Closing the Loop}
            Given the solution to Eq.~\eqref{eq:exp_learning_opt}, the following step is to execute the behaviors on the robot, record the measured data and repeat the process until a desired parameter confidence defined by $p(\theta | \mathcal{D})$.
            Since our problem formulation only accounts for the posterior parameter mode $\hat{\theta}$, we use the approximate update equation~\cite{schmidt2014dart} on the posterior variance
            \begin{equation}
            \label{eq:bayesian_paramvar_update}
                \Sigma_\theta^{+} = \left(\mathcal{F}(\mathcal{D} | \tau, \hat{\theta}) +  \Sigma_\theta^{-1} \right)^{-1}
            \end{equation}
            where the Fisher information is calculated from the collected experiment data, $\Sigma_\theta$ is the covariance of the prior, $\Sigma_\theta^+$ is the posterior covariance, and $\hat{\theta}$ is solved initially using Eq.~\ref{eq:contact_aware_max_like}.
            Note that this equation over-approximates the uncertainty, which we found in practice assists with parameter convergence.
            \begin{remark}
                Because we are estimating continuous parameters $\theta$, and not contact forces or potentially discontinuous state transitions, we can assume belief propagation is also continuous. Thus, we do not require techniques such as saltation matrices \cite{Kong_2024} to estimate parameter belief.
            \end{remark}

\begin{figure}[h!]
        \centering
        \includegraphics[width=\linewidth]{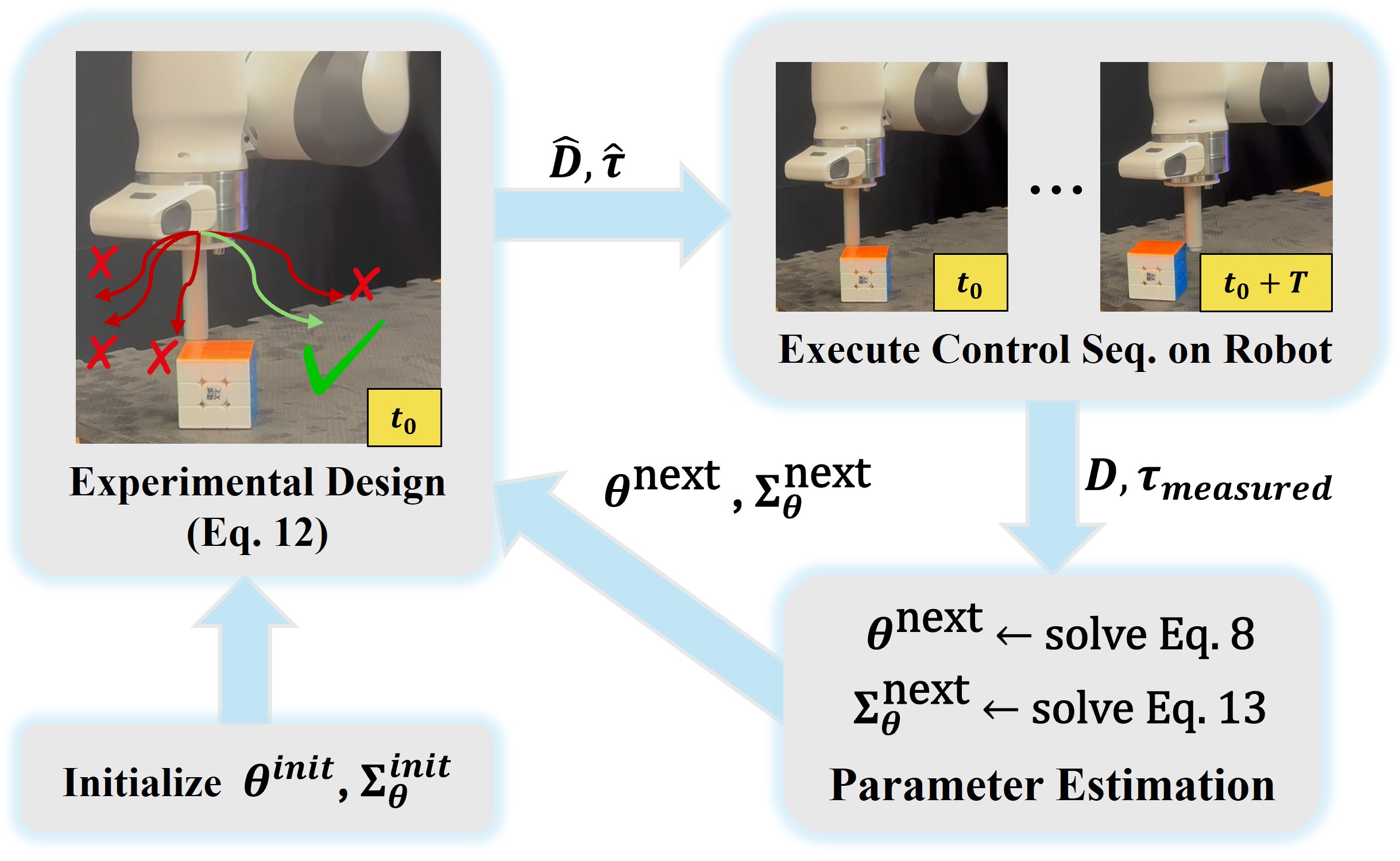}
        \caption{\textbf{Experimental Design Pipeline. } Here we outline our approach using Alg. \ref{alg:exp_des}. Given a prior on the parameter, we plan trajectories that excite information-rich contact modes. We then apply the planned trajectories to the robot and gather sensor measurements. These measurements are used to update the parameters and the variance of the prior to obtain the posterior. We repeat this process until the robot estimates parameters at a parameter certainty tolerance of choice.}
        \label{fig:pipeline}
    \end{figure}

    We present an algorithm that comprises of 1) the contact-aware experimental design optimization that searches for information-rich contact, and 2) the contact-aware maximum a-posteriori estimation problem that estimates parameters from collected data in Algorithm~\ref{alg:exp_des}. 
    We summarize the contact-aware experimental design procedure in Fig.~\ref{fig:pipeline}.

        \begin{algorithm}
        \caption{Contact-Aware Behavior Synthesis}\label{alg:exp_des}
            \begin{algorithmic}[1]
            \Require initial state $x_0$, prior parameter estimate $p(\theta)$, parameterized dynamics, sensor, and contact models, total number of experiments $k_\text{max}$, and experiment duration $T$.
            % nominal plan $\Pi(u | x_0, \theta)$, total time iterations $T$, time horizon $t_H$, time step $dt$
            \State $k = 0$
            \While{$k< k_\text{max}$}
                \State $\hat{\mathcal{D}}, \hat{\tau} \gets$ CA-OED Eq.~\eqref{eq:exp_learning_opt} \Comment{calc. contact behavior}
                \State $\mathcal{D}\gets$ Execute experiment $\hat{\mathcal{D}}, \hat{\tau}$ on robot
                \State $\hat{\theta}, \Sigma_\theta^+ \gets$ CA-MAP Belief update Eq.~\eqref{eq:contact_aware_max_like}, Eq.~\eqref{eq:bayesian_paramvar_update}
                \State $k \gets k +1$
                % \State $\{u_{t=1}^{t+t_H} \}_{j=1}^M \thicksim \Pi(u | x_0, \theta)$ \Comment{Sample controls}
                % \State $ x^{t+dt} \gets \min_{x,u} \mathcal{J}_\mathcal{I}(x_0, \{u_{t=1}^{t+t_H} \}_j, \theta)$ \Comment{min. cost}
                % \State $\Pi(x_0,\theta) \gets \Pi(x^{t+dt},\theta)$\Comment{update plan}
                % \State $x_0 \gets x^{t+dt}$ \Comment{update $x_0$}
            \EndWhile
            \end{algorithmic}
        \end{algorithm}

\begin{figure*}[t!]
        \centering
        \includegraphics[width=0.95\textwidth]{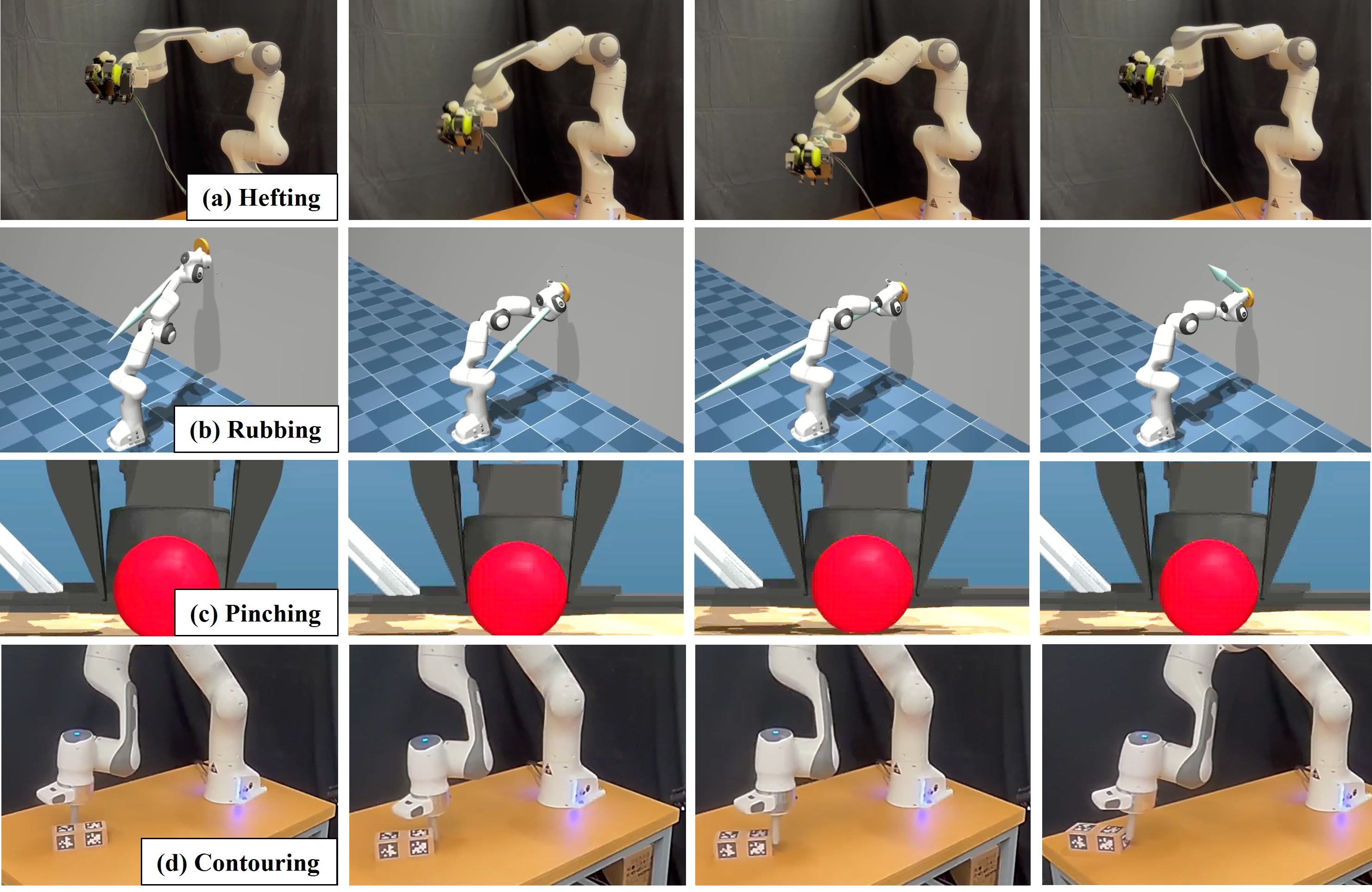}
        \caption{\textbf{Emergent Behaviors from Experimental Design. } Time series of experiments conducted over all scenarios. By maximizing contact-aware Fisher information, the robot seeks information-rich contacts for efficient identification of (a) mass, (b) friction coefficient, (c) material properties, and (d) shape. Simulation results are visualized on MuJoCo \cite{todorov2012mujoco}. The robots were given a planned trajectory to execute, where we assume perfect robot state tracking upon executing the trajectory. We see the emergent behaviors in (a) a hefting motion that perturbs contact force on the object, (b) the end-effector rubbing against the wall with a nonzero normal force and tangent contact velocity, (c) the gripper pinching the object to excite contact modes, and (d) an object contouring behavior with the end effector to excite contact modes across different regions of the object.}
        \label{fig:results_emerging_behaviors}
\end{figure*}

\begin{figure*}[t]
  \centering
  \subfloat[\textbf{Mass Estimation. } The robot seeks to estimate the mass of a grasped object.]{\includegraphics[width=0.49\linewidth]{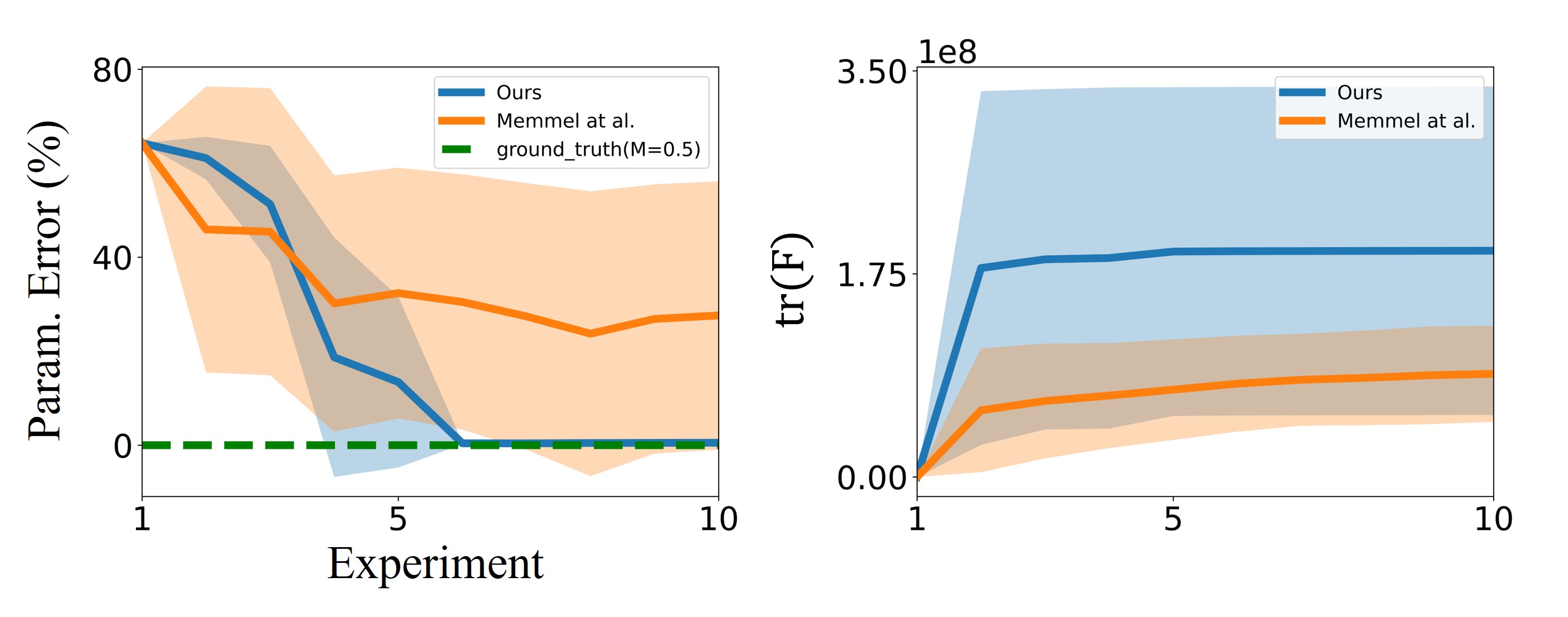}\label{fig:heft_res}}
  \hfill
  \subfloat[\textbf{Friction Estimation. } The robot seeks contacts with the wall to estimate the friction coefficient.]{\includegraphics[width=0.49\linewidth]{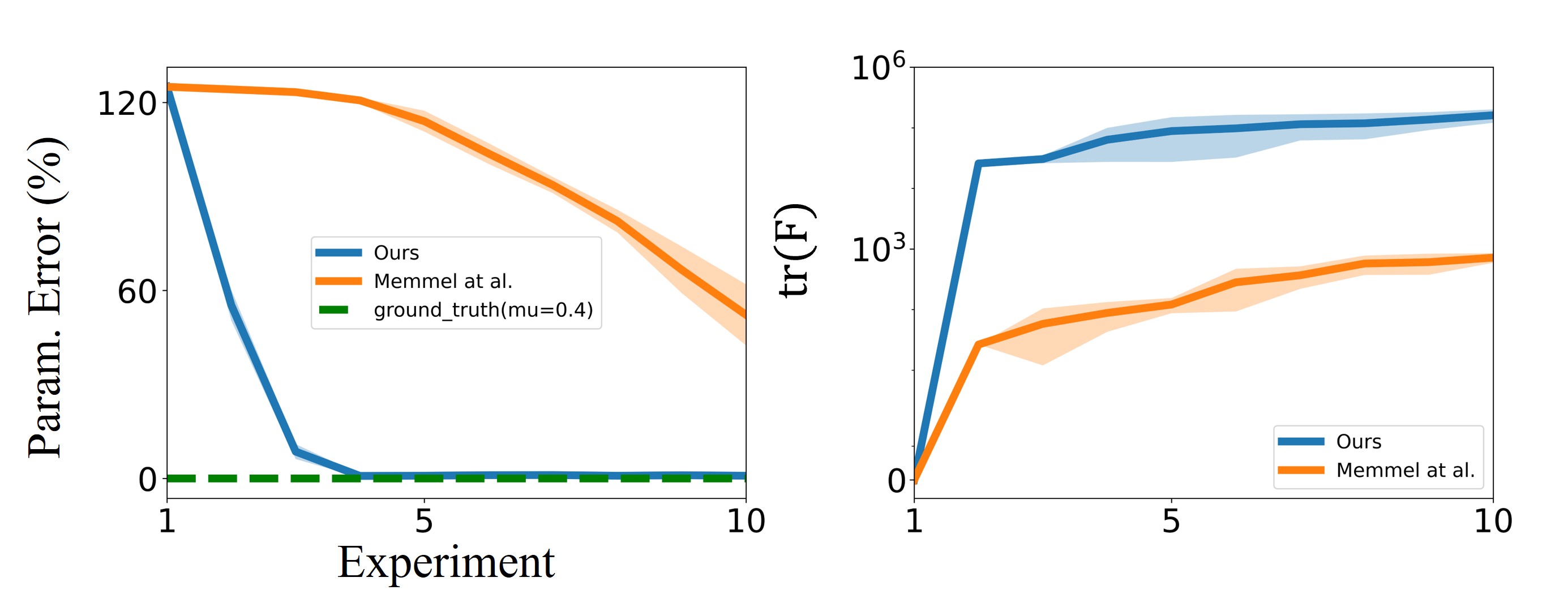}\label{fig:reb_res}}\\[1ex]
  \subfloat[\textbf{Material Property Estimation. } The robot gripper interacts with an object to learn its material properties (stiffness and dampening coefficients).]{\includegraphics[width=0.49\linewidth]{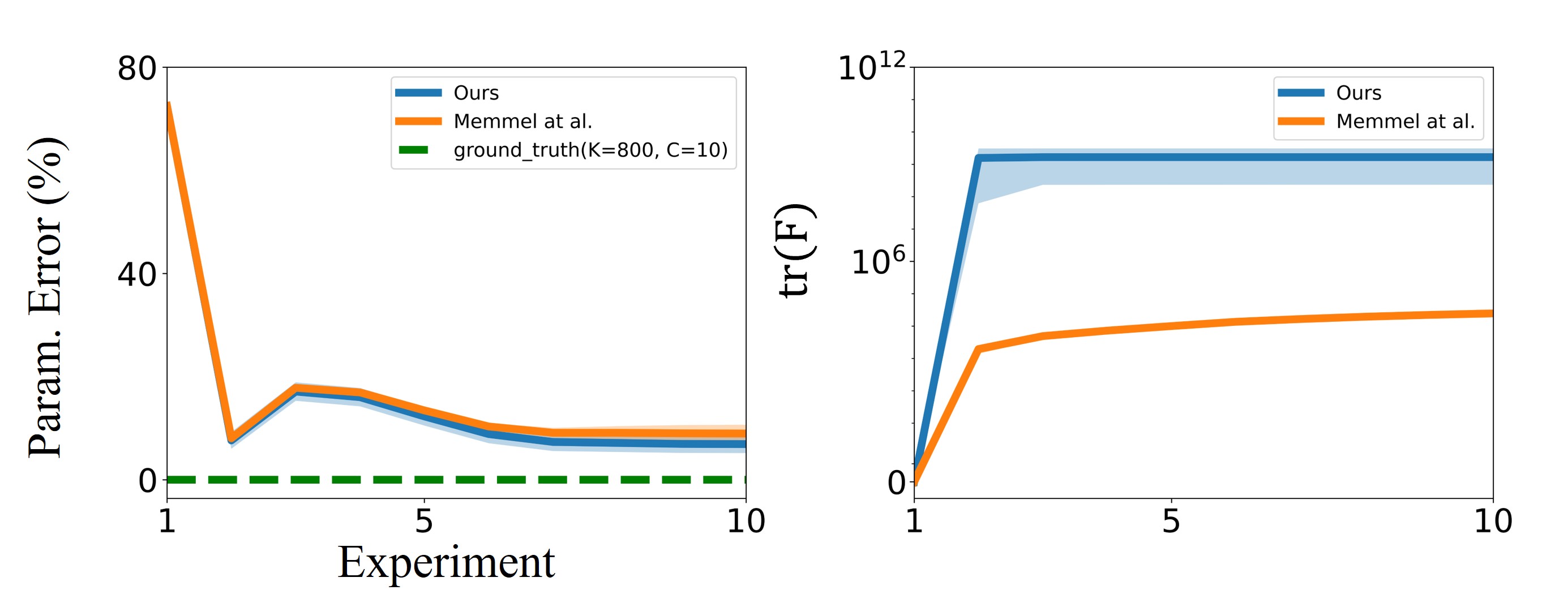}\label{fig:pinch_res}}
  \hfill
  \subfloat[\textbf{Shape Estimation. } The robot that interacts with a box to learn its shape (length and width). Estimation is solved in-the-loop ober discrete time.]{\includegraphics[width=0.49\linewidth]{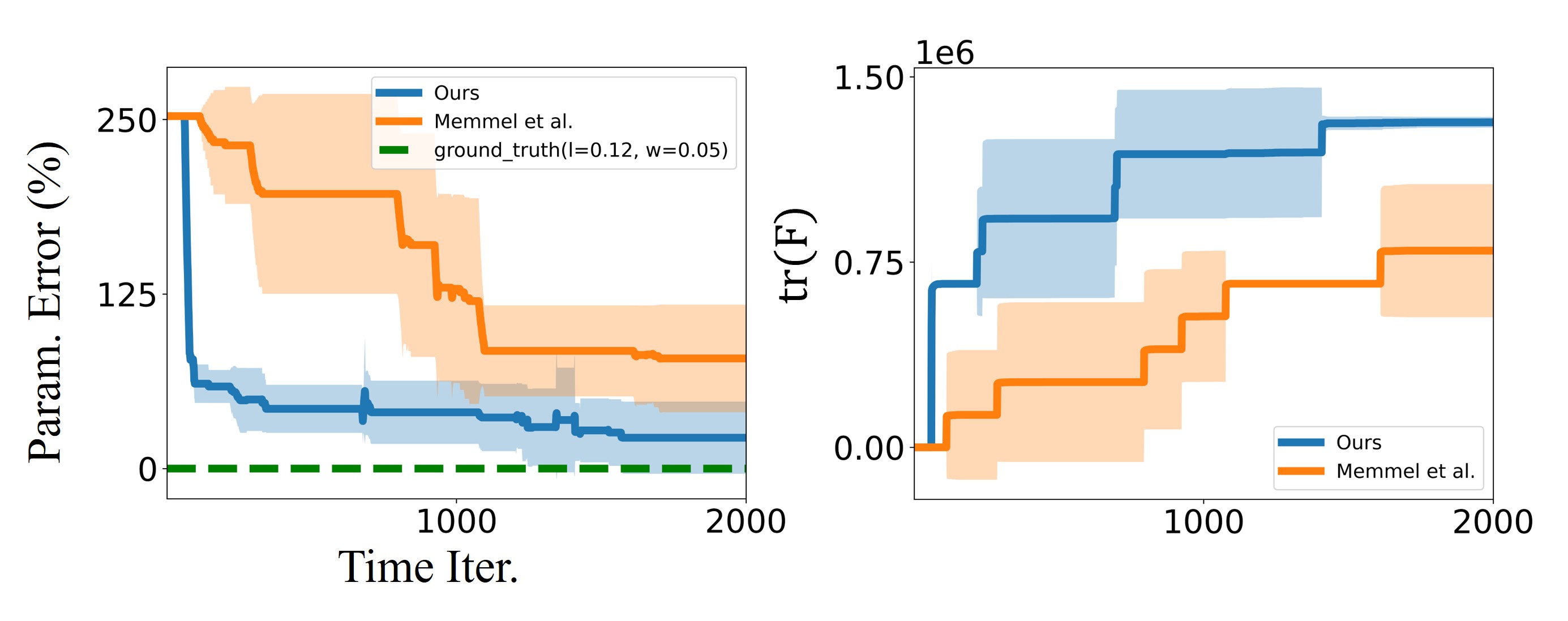}\label{fig:contour_res}}
  \caption{\textbf{Contact Excitations Improves Parameter Learning. } We compare our experimental design approach with a baseline similar to works in \cite{memmel2024asidactiveexplorationidentification} for all robot scenarios. The baseline implements a belief-space planning controller that reasons about contact na\"ively. We evaluate each robot scenario over a finite number of experiments. Here, we model the parameter error as $\%\mathrm{err} = 100 (\hat{\theta} - \theta^*)/\theta^*$, where $\hat{\theta}$ is the current estimate and $\theta^*$ is the true parameter. By maximizing the contact-aware Fisher information, we see that the robot seeks contacts that are information-rich about the unknown parameters of interest, and thus converging closer to the true parameter values $\theta^*$, with resulting higher Fisher information gains.}
  \label{fig:results}
\end{figure*}

\section{Results and Discussion}
\label{sec:results_and_discussion}
        
    This section details simulated and experimental results of our contact-aware experimental design approach in several scenarios. Our approach is evaluated using contact-implicit trajectory optimization methods where we optimize for contact modes that maximize information about objects with unknown parameters of interest, thus improving parameter learning performance. We compare our approach to a similar approach~\cite{memmel2024asidactiveexplorationidentification}, which executes trajectories based on a Fisher information maximization approach where contact is not considered explicitly. Particularly, we are interested in observing emergent behaviors from our approach that answer the following questions:

    \begin{enumerate}[label=Q{\arabic*})]
        \item What classes of parameter learning problems require more \emph{compounding} and \emph{constant} contact interactions as the resulting optimal emergent behavior?
        \item What classes of parameter learning problems require \emph{making} and \emph{breaking} contact as its resulting optimal emergent behavior?
        \item What convergence do we see for parameter learning through our optimized contact behaviors? 
        \item How does the information loss landscape change for different parameter estimation problems?
        % \item What is the composition of the contact data-set that we produce?
    \end{enumerate}
    We answer the questions aforementioned in this section sequentially with corresponding answers `A(-)'.

    \noindent\underline{Robot Scenarios}
    
    The results in this paper answers these questions by implementing our Fisher information maximization approach on the following robot scenarios shown in Fig. \ref{fig:exp_setup}:

    \begin{enumerate}
        \item \textbf{Weight Estimation}, where an Allegro-Franka system estimates the mass of a grasped object through hefting behaviors. 
        \item \textbf{Friction Estimation}, where a Franka robot interacts with a wall to estimate friction coefficient.
        \item \textbf{Stiffness Estimation}, where a robot gripper learns material stiffness and dampening properties of a uniform object through its grasping interactions.
        \item \textbf{Shape Estimation}, where a Franka robot estimates the shape of a box (length and width) through non-prehensile contact interactions.
    \end{enumerate} 

    Experiment details are shown in Appendix \ref{app:exp_details_appendix}. Note that we implement these robot scenarios in the presence of \emph{high contact force sensor noise}, which makes small perturbations of contact ambiguous information to identify parameters.
    We demonstrate the robustness of our approach to high sensor noise by enabling robots to reason about exploring for information-rich contacts \emph{outside} some contact uncertainty envelope. 
    
    Our experimental pipeline is shown in Fig. \ref{fig:pipeline}. 
    We compare our approach to a recent belief space planning baseline in \cite{memmel2024asidactiveexplorationidentification}, which explores for information-rich data about parameters without the explicit reasoning over contact.
    % We compare our approach to random-sampling baselines, where the robot randomly samples (from a uniform distribution) trajectory waypoints to collect sensor data for parameter learning.

    \noindent\textbf{A1. Compounding Contact Behavior}

    As shown in results in Fig. \ref{fig:results_emerging_behaviors}, the hefting experiment (a), where the robot pregrasps an object and learns the mass of that object through perturbing its constant normal contact force, as well as the rubbing experiment (b), where the robot interacts with a nearby wall to estimate the friction coefficient, both maintain nonzero contact in order to estimate parameters.
    When estimating friction coefficient, we can see in the contact sensor model in Appendix \ref{app:contact_sensor_model} that this term is identifiable when exciting both a high normal force as well as a non-zero tangent velocity. Therefore, maximizing over contact-aware Fisher information yields a high excitation of tangent velocities at the contact frame along with a high excitation of normal contact force. Therefore, the resulting emergent behavior results in compounding frictional contact with the wall.

    In the case of the hefting experiments (a), it is possible to estimate mass parameters \emph{even if} the object \emph{momentarily} leaves the hand of the robot as long as the object returns to the hand. However, this adds new sparsities in the contact forces, and thus the sensor readings during the moments where the object leaves the hand would not contribute to significant updates of the parameter prior. Therefore, in order for the robot to \emph{effectively} learn the mass parameter of the object, it is desirable for the object to not leave the hand, and rather experience \emph{perturbations} in normal contact forces. Such perturbations would excite the contact sensor model output written in Appendix \ref{app:contact_sensor_model} such that the gradient of the sensor model can extract mass information from the signed distance function $\phi_n$ between the object and the hand, as well as the normal velocity at the contact frame $v_n$. As a result, the resulting behavior for this experiment using our experimental design approach yields a hefting behavior where the robot moves its hand in an up-and-down motion in order to perturb normal force readings between the object and the robot \emph{optimally}.

    \noindent\textbf{A2. Making and Breaking Contact Behavior}
    
    As shown in Fig. \ref{fig:results_emerging_behaviors}, the pressing experiment (c), where a robot gripper interacts with an object by grasping it, as well as the contouring experiment (d), where the robot interacts with an object its end effector, both make and break contact in order to significantly perturb the contact forces. Note that for experiment (d), the robot identifies the parameters online, where we execute Alg.~\ref{alg:exp_des} in the loop during robot operation. In these experiments, the robot is tasked with reasoning about material properties stiffness and dampening (c), as well as shape (d). To do so, our experimental design approach gives us a planned trajectory that excites contact modes with the object by making and breaking contact in order to significantly perturb the contact forces. Intuitively, learning material properties naturally requires excitations in the normal forces between the robot and the object, and in doing so we can excite gradients in the contact sensor model as written in Appendix \ref{app:contact_sensor_model}.

    \noindent\textbf{A3. Parameter Convergence Over Experiments}

    We aim to demonstrate the convergence of parameter learning by comparing our experimental design approach for all robot scenarios shown in Fig. \ref{fig:results_emerging_behaviors} to a baseline similarly implemented in \cite{memmel2024asidactiveexplorationidentification}.
    The baseline reasons about information-rich trajectories to improve learning, only that it does not reason over contact explicitly by simply approximating the gradients of the sensor model using a finite differencing method.
    The goal of this section is to show that by reasoning over contact explicitly, we can converge better on parameter uncertainty. 

    The results are shown in Fig. \ref{fig:results}. 
    We can see that upon evaluating over experimental runs, we can observe that our experimental design approach is able to converge to a more precise parameter estimate with higher Fisher information gains. 
    The emergent behaviors that result from contact-aware Fisher information maximization actively search for contact modes that are information-rich, resulting in naturally emerging behaviors that allow for precise identification of parameters.
    The reason that our approach improves the works in~\cite{memmel2024asidactiveexplorationidentification} is due to our direct consideration of contact interactions in the optimization as well as our approach in structuring the Fisher information with contact. 
    The baseline approaches contact more na\"ively, where Fisher information is constructed by approximating the gradients of the trajectory measurement model through a finite differencing method for a small perturbation in parameter space. 
    By structuring the Fisher information based on a structured contact measurement model, the robot is able make deliberate decisions through its contact seeking behaviors.

    In the case of the pinching experiments in Fig.~\ref{fig:results_emerging_behaviors}, we can observe that our contact-aware Fisher information maximization approach only marginally improves parameter learning compared to the baseline. This due to the constrained nature of the problem, where the robot motion is constrained by a bounded 1 dimensional motion, assuming that the gripper is at a pre-grasped state. Because of high contact sensor noise, the robot is forced to perturb the contact force between its grippers and the object such that the resulting contact readings are outside of the contact uncertainty envelope in order to properly identify parameters. Because of the highly constrained nature of contact in this robot scenario, directly reasoning over information-rich contact will result in similar emergent behaviors as the baseline approach. Similar behaviors for the hefting experiments can be seen if the contact sensor is not as noisy. However, we can observe that our approach can outperform the baseline in more complex and unconstrained scenarios.

\begin{figure}
        \centering
        \includegraphics[width=\linewidth]{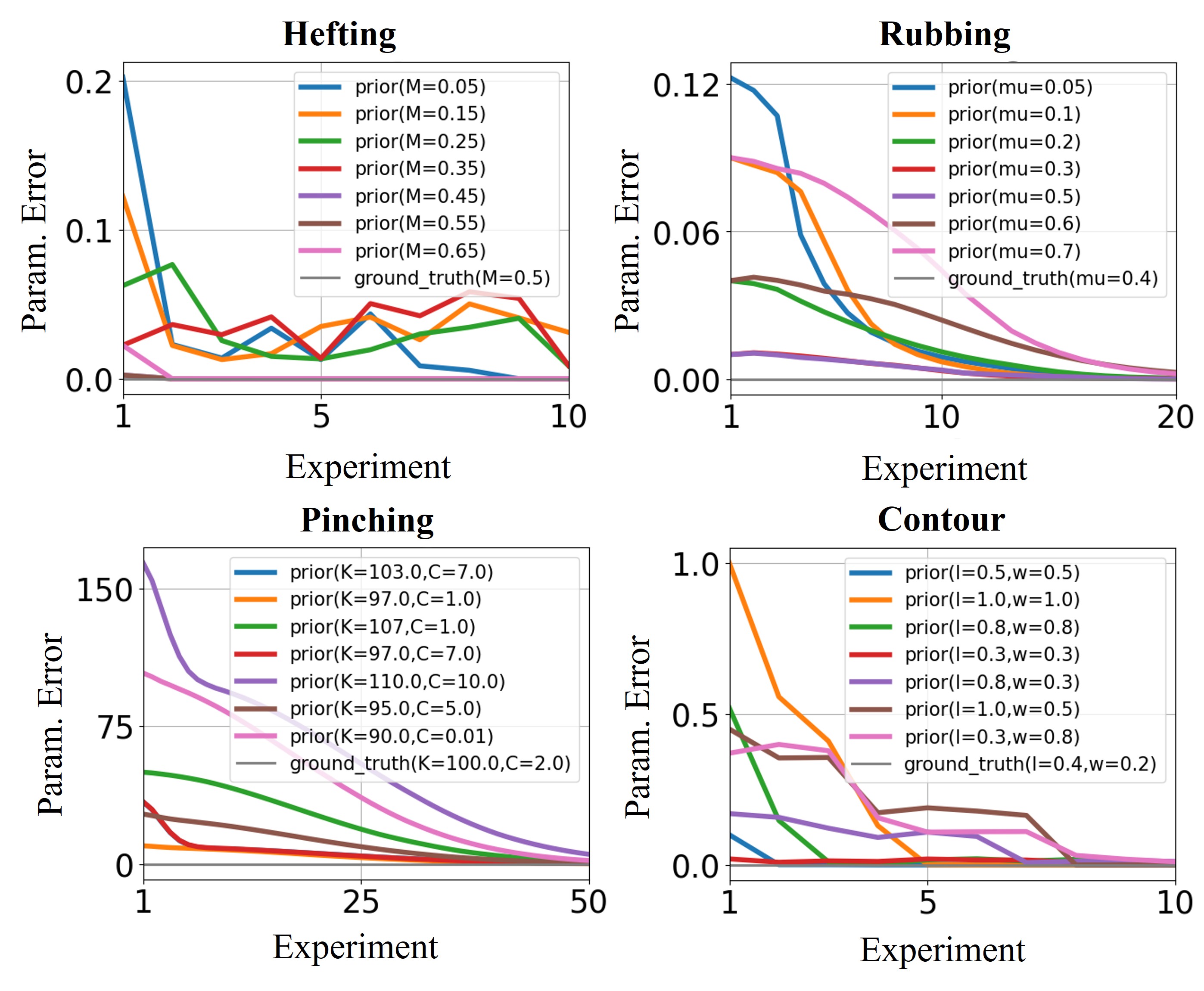}
        \caption{\textbf{Robustness to Initialization. }We study robustness of our approach is to 7 initial parameter priors for all scenarios. Note here that we model the parameter error as a distance $\delta = |\hat{\theta} - \theta^*|$. We observe that our contact-aware Fisher information maximization approach is resilient across a diverse initialization of priors over all experiments.}
        \label{fig:robustness}
    \end{figure}

    In Fig.\ref{fig:robustness}, we demonstrate the robustness of our approach as a function of parameter initialization.
    We observe that our approach is highly robust to initial priors within $\sim$30\% for hefting, and even more robust for other scenarios. 
    We also expand on the remarks in \cite{nima} that estimating multiple parameters that are \emph{co-dependent} to one another makes the estimation unidentifiable, unless we make assumptions of the contact interactions (e.g. bounding contact impulses), provided within the problem definition in Eq. 8 in the paper.

    \begin{figure*}[t!]
        \centering
        \includegraphics[width=0.9\textwidth]{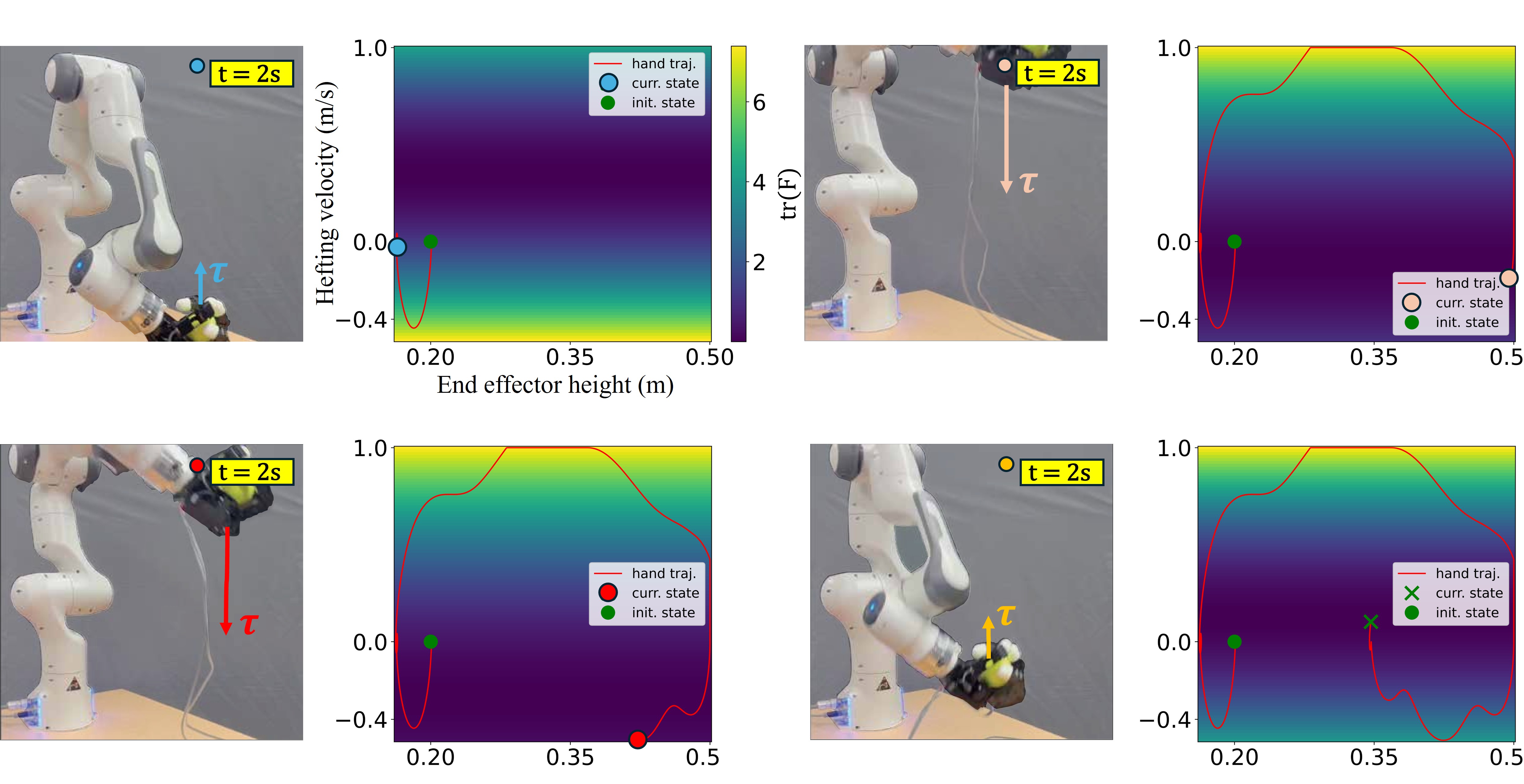}
        \caption{\textbf{Hefting Behavior Information Landscape. } The robot excites penetration deformation $\phi_n$ by varying its hand positions and non-zero contact normal velocity $v_n$ in order to excite information-rich contact about mass, resulting in an up-and-down hefting behavior shown in Fig. \ref{fig:results_emerging_behaviors}}
        \label{fig:heft_inf}
    \end{figure*}

    \begin{figure*}[t!]
        \centering
      % \subfloat[\textbf{Hefting Behavior Information Landscape. } The robot excites penetration deformation $\phi_n$ by varying its hand positions and non-zero contact normal velocity $v_n$ in order to excite information-rich contact about mass, resulting in an up-and-down hefting behavior shown in Fig. \ref{fig:results_emerging_behaviors}]{\includegraphics[width=0.9\textwidth]{figs/inf_landscape/FIG-heft_inf_landscape_v2.jpg}\label{fig:heft_inf}}\\[1ex]
      % \vspace{-0.5em}
      \subfloat[\textbf{Pinching Behavior Information Landscape. } Here, Fisher information is maximized by exciting contacts by activating the penetration distance into the object $\phi_n$ as well as the deformation rate $\frac{\partial \phi_n}{\partial t}$ through pinching. This yields natural behaviors of squeezing and contracting the object in order to identify material properties optimally.]{\includegraphics[width=0.9\textwidth]{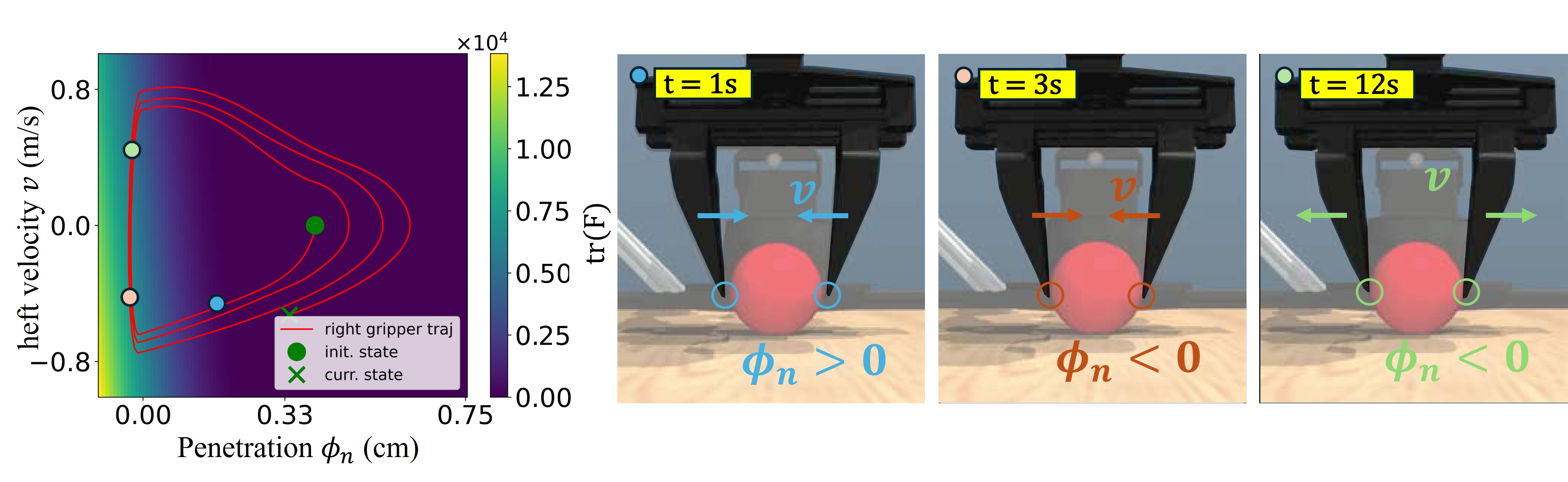}\label{fig:pinch_inf}}\\[1ex] 
      \vspace{-0.5em}
      \subfloat[\textbf{Rubbing Behavior Information Landscape. } Here, information is maximized by exciting normal contact velocity $v_n$ and tangent contact velocity $v_t$ in order to collect information-rich contact about the friction coefficient, resulting in a rubbing behavior shown in Fig. \ref{fig:results_emerging_behaviors}]{\includegraphics[width=0.9\textwidth]{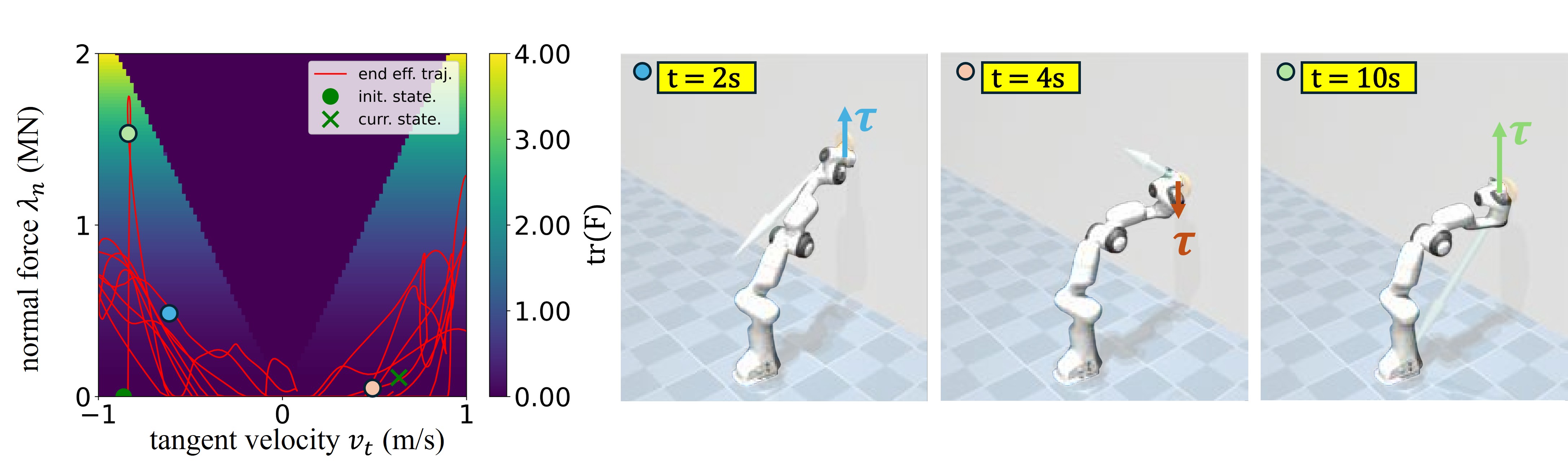}\label{fig:rub_inf}}
      \caption{\textbf{Optimizing Over Contact-Seeking Information Landscapes. } We show the contact-aware Fisher information landscape that our experimental design method optimizes over for the stiffness/dampening and friction estimation scenarios.}
      \label{fig:inf_landscape}
    \end{figure*}

    \noindent\textbf{A4. Information Loss Landscape}

        We aim to provide a visual intuition of the information landscape that we optimize over in our contact-aware Fisher information maximization problem in Eq. \ref{eq:exp_learning_opt}. The information landscape is dictated by the physics models chosen for each robot scenario as well as the unknown parameters to estimate. Given that the robot has engaged contact with the environment, we show in Fig. \ref{fig:inf_landscape} the changes in the information landscape  with respect to different kinematic parameters that impact the landscape the most.

        Based on the results in Fig. \ref{fig:inf_landscape}, we choose to focus on observations from the information landscape for the weight, friction, and material stiffness estimation scenarios.
        
        \noindent\underline{Weight Estimation} 
        
        The emergent hefting behaviors shown in Fig. \ref{fig:results_emerging_behaviors} optimizes for perturbations in normal contact between the hand and the object, with assumed soft body interactions. As a result, a natural consequence from our Fisher information maximization approach is exciting the `perturbation' depth $\phi_n$ between the hand and the object upon contact and the normal velocity $v_n$ at the contact frame. Fig. \ref{fig:heft_inf} shows the smooth information landscapes resulting from the current configuration of the hand-object system. The Fisher information landscape is linear with respect to the kinematic inputs, and results in the system being `pulled' to the velocity boundaries, resulting in the robot exhibiting a hefting behavior.

        \noindent\underline{Stiffness Estimation}
        
        The results in Fig. \ref{fig:results_emerging_behaviors} show that the robot optimally learns stiffness and dampening parameters of a ball by interacting with its gripper. 
        The gripper excites perturbations in the normal forces between the gripper fingers and the ball by making and breaking contact, and pressing into the ball during contact engagement. 
        The information landscape shown in Fig. \ref{fig:pinch_inf} indicates that there is a locally optimal trade-off between the penetration $\phi_n$ length of the gripper finger and the ball surface and the penetration rate $\frac{\partial \phi_n}{\partial t}$ (equivalent to normal contact velocity $v_n$). 
        We can observe that such normal force excitations can lead to excitations of the gradients of the sensor model detailed in Appendix \ref{app:contact_sensor_model}. 
        The robot excites its grippers to pinch and release contact with the object by seeking trajectories that maximize penetration and pinching velocity as shown by the information landscape.
        This naturally results in the robot exhibiting a `squeezing and contracting' behavior with the object. 

        \noindent\underline{Friction Estimation}

        From our approach, we observe in Fig. \ref{fig:results_emerging_behaviors} that the robot optimally learns the friction coefficient between its end-effector and the wall by maintaining non-zero tangent contact forces through rubbing. 
        Intuitively, the friction coefficient parameter appears only in the tangent forces in the sensor model used for the experiment (See Appendix \ref{app:contact_sensor_model}), so the robot requires consistent and compounding excitations of tangent forces by pressing against the wall with a normal contact velocity $v_n$ and exerting a non-zero tangent velocity $v_t$ at the contact frame. 
        
        As shown in Fig. \ref{fig:rub_inf}, we visualize the information landscape that the robot optimizes over to obtain information-rich data to estimate the friction coefficient. 
        The robot maximizes information about the friction coefficient  during contact by sliding velocity $v_t$ and pressing velocity $v_n$ against the wall simultaneously, thus exhibiting a rubbing behavior. 
        Note in the figure that the robot seeks to choose velocities at contact that are approach the peaks of the information landscape, but are bounded by physical constraints. 
        This yields the robot in exhibiting a rubbing behavior with the wall.
    
\section{Limitations}
\label{sec:lims}

    The experimental design method in this paper has several limitations worth discussing below.

    \textbf{Deterministic Dynamics in a Dynamically Changing World.} Our approach assumes that the dynamics of the system are deterministic, and we fit stochastic data with a sensor model that assumes zero mean Gaussian noise. Such assumptions can be especially tricky to deal with since we do not capture any uncertainties in our dynamics model, which may lead to under or over approximations of contact behaviors and limit the performance of our method in practice. Additionally, our dynamics model captures contact modes at certain body frames of the robot system; if the robot interacts with the world outside the body frame of interest, the desired synthesized behavior may not be accurately exhibited on the robot.
    
    \textbf{Nonlinear Optimization Landscape.} In general, the complexity of the sensor model and dynamics of the robot create a challenging optimization landscape in the CA-MAP problem in Eq. \ref{eq:contact_aware_max_like} that is nonlinear and non-convex in nature. As a result, optimizing contact behaviors for parameter learning are likely to be sub-optimal solutions. Consequently, finding effective behaviors in real-time settings can be difficult to plan for.

    \textbf{Contact Model Parameter Tuning.} The experiments presented in this paper requires a great deal of parameter tuning in the contact model. For example, the hefting experiments demonstrated in this paper assumes that we have a priori knowledge of the material properties of the object of interest, which is required for the soft contact model implemented. This required careful tuning of the contact model parameters in order to obtain reasonably realistic contact model behaviors.

    \textbf{Limitations from Fisher Information.} Optimizing over Fisher information, in accordance with Definition \ref{def:CRLB}, reduces the variance of the parameter estimates by pushing the tight lower bound down. Future work will investigate stronger notions of variance reduction by means of reducing an \emph{upper bound} of parameter variance. Additionally, Fisher information only provides notions of parameter \emph{precision}, and no guarantees on accuracy, which therefore may lead to systematic error in parameter estimates over time. 

\section{Conclusion and Future Work} 
\label{sec:conclusion}

This paper presents a contact-aware Fisher information maximization approach that results in emergent learning behaviors that precisely identifies parameters of interest. We develop the theoretical foundations of our problem formulation, and why maximizing over Fisher information yields the steepest descent in parameter learning, thus improves convergence of parameters through select contact behaviors. Last, we demonstrate our approaches ability to estimate physics parameters on a variety of robot settings, resulting in different emergent learning behaviors depending on the parameters being estimated. 

Future work will explore how our Fisher information maximization approach will impact downstream policy learning and trajectory optimization. A limitation of the proposed work is that our approach structures learning using deterministic dynamics. Future work will explore how process noise impacts our contact-aware information maximization approach. Furthermore, future directions will also explore contact-aware information maximization approaches to more complex, higher dimensional robotic systems.
%% Use plainnat to work nicely with natbib. 
% \section{RSS citations}

% Please make sure to include \verb!natbib.sty! and to use the
% \verb!plainnat.bst! bibliography style. \verb!natbib! provides additional
% citation commands, most usefully \verb! \citet!. For example, rather than the
% awkward construction 

\section{Acknowledgements}
    This work is supported by the National Science Foundation under award NSF FRR 2238066. Any opinions, findings, and conclusions or recommendations expressed in this material are those of the authors and do not necessarily reflect the views of the National Science Foundation. 

    The authors extend their gratitude to Ethan K. Gordon and Michael Posa from the GRASP Lab at the University of Pennsylvania for all the insightful discussions and constructive feedback for this work.
    
\bibliographystyle{plainnat}
\bibliography{main}

\appendices
\section{Proofs}
    \subsection*{Fisher Information Steepest Ascent Proof}
    \label{app:steepest_asc_proof}
        \begin{proof}
            We first approximate $\mathcal{L}(\mathcal{D} | \tau, \theta + \delta\theta)$ in Eq.\ref{eq:steepest_asc_problem} by a second order Taylor expansion,

        \begin{equation} 
        \label{eq:L_taylor_approx}
            \begin{split}
               \mathcal{L}(\mathcal{D} | \tau, \theta + \delta\theta) &\approx \mathcal{L}(\mathcal{D} | \tau, \theta) + \nabla_\theta \mathcal{L}(\mathcal{D} | \tau, \theta)^T \delta\theta \\ &+ \frac{1}{2}\delta\theta^T \nabla_\theta^2 \mathcal{L}(\mathcal{D} | \tau, \theta) \delta\theta 
            \end{split} 
        \end{equation}
        % By observing the Taylor expansion, we can see that the Fisher information appears with the Hessian term, since $\mathcal{F}(\mathcal{D} | \tau, \theta) \approx -\nabla_\theta^2 \mathcal{L}(\mathcal{D} | \tau, \theta)$, at the expectation, as per Eq. \ref{eq:general_fisherinf}. Substituting Eq. \ref{eq:L_taylor_approx} into Eq. \ref{eq:steepest_asc_problem} yields
        Solving for $\delta\theta$ can then be done with the Taylor Expansion as

        \begin{equation}
            \delta \theta^\star = \argmax_{\delta \theta} \nabla_\theta \mathcal{L}(\mathcal{D} | \tau, \theta)^T \delta\theta + \frac{1}{2}\delta\theta^T \nabla_\theta^2 \mathcal{L}(\mathcal{D} | \tau, \theta) \delta\theta
        \end{equation}
        which has the following closed form solution
        
        \begin{equation}
            \delta\theta^* = \mathcal{F}(\mathcal{D} | \tau, \theta)^{-1} \nabla_\theta \mathcal{L}(\mathcal{D} | \tau, \theta).
        \end{equation}
        with $\mathcal{F}(\mathcal{D} | \tau, \theta) = - \nabla_\theta^2 \mathcal{L}(\mathcal{D}|\tau, \theta)$ being the Fisher information matrix~\cite{pmlr-v70-sun17b}.
        % Note that in the steepest descent solution, we assume that $\mathcal{F}$ is an invertible matrix, which is challenging to perform since $\mathcal{F}$ is often ill-conditioned in practice \cite{cintrónarias2020sensitivitymatrixbasedmethodology,890346}. However, by performing experimental design using Eq. \ref{eq:exp_learning_opt}, we improve the condition number $\kappa(\mathcal{F})$, and by searching for parameters $\theta$ simultaneously, we regularize the spectrum $\mathrm{spec}(\mathcal{F})$ such that $\kappa(\mathcal{F}) \approx 1$. Performing an inversion over the whole matrix (rather than a subspace of the information landscape \cite{890346}) is advantageous when computing steepest descent for proper identification of all parameters.
        \end{proof}

\section{Implementation Details}

    \subsection*{Relationship between Condition Number and Trace} 
    To improve the condition number of the Fisher information matrix, $\kappa(\mathcal{F})$, we optimize over the trace of the Fisher information matrix (T-optimality) as condition numbers bounded by the trace of a nonsingular matrix~\cite{FENNER1974157}, as
    \begin{equation}
        \begin{split}
            \left(||\mathcal{F}||^2||\mathcal{F}^{-1}||^2 -\left(n-2\right) \right)^2 \geq \left(\kappa\left(\mathcal{F}\right) + \kappa^{-1}\left(\mathcal{F}\right)\right)^2 &\\
            \geq \left\{
            \begin{array}{c}
            n \ \mathrm{even}, \frac{4}{n^2} ||\mathcal{F}||^2||\mathcal{F}^{-1}||^2 \\
            n \ \mathrm{odd}, \frac{4}{n^2 - 1} \left(||\mathcal{F}||^2||\mathcal{F}^{-1}||^2 -1 \right)
            \end{array}
            \right. 
        \end{split}
        \end{equation}
        where $\mathcal{F} \in \mathbb{R}^{n \times n}$, and the Schur matrix norm $||\mathcal{F}|| = \mathrm{tr}\left(\mathcal{F}^T\mathcal{F}\right)$, for a nonsingular matrix $\mathcal{F}$. By maximizing over the trace of the Fisher information matrix in Eq.~\ref{eq:exp_design_formulation}, we thus maximize the bounds of the condition number.
    
    % \subsection*{Parameter Variance Updating}\label{subsec:param_var_appendix}

    %     In reference to the solution to solving unknown parameters $\theta^*$ via maximum likelihood for a linear regression model, we can observe the following inverse expression term that acts as an update step in our parameter variance,

    %     \begin{equation}
    %         \Sigma_{\theta,post.} = \left( \mathcal{F}(\mathcal{D} | \theta) + \Sigma_{\theta,prior}^{-1} \right)^{-1}
    %     \end{equation}
    %     Additionally, this implicitly constrains the variance update to the information-richness of our data using the Fisher information.

    \subsection*{Contact Model}
    \label{app:contact_sensor_model}

        The contact sensor model implemented in this paper utilizes the following soft contact model,
        \begin{equation}
        \label{eq:soft_contact_model}
            \begin{bmatrix}
            \lambda_n \\
            \lambda_t
            \end{bmatrix} = 
            \begin{bmatrix}
                \max \left( 0, -K \phi_n - C |v_n|\right) \\
                \hat{v_t} \max \left( -\mu \lambda_n, -R |v_t| \right)
            \end{bmatrix}
        \end{equation}
        where $K$, $C$, $\mu$, and $R$ are the material stiffness, material dampening, friction coefficient, and friction resistance parameter, respectively, $\phi_n$ is the signed distance function between the system body and object, and $v_c = [v_n, v_t]^T = \mathbf{J_c}(q) \dot{q}$ are the normal and tangent velocities at the contact frame with unit vectors $\hat{v}_n$ and $\hat{v}_t$, respectively, and where $\mathbf{J_c}(q)$ is the contact Jacobian that maps joint velocities to the contact frame.

    \subsection*{Experiment Details}
    \label{app:exp_details_appendix}

        Here we outline the experiment settings in this paper.

        \noindent\textbf{Predictive Sampling Parameters. } All experimental setups were conducted in discrete time settings with a short time horizon $t_H = 10$, and an initial nominal plan $\Pi_{\mathrm{init}} = \mathrm{zeros}(t_h,n_a)$, where $n_a$ is the number of controllable joints in the system. The $N = 10$ sampled control splines were constructed using a Cubic Hermite spline, as demonstrated by~\cite{howell2022predictivesamplingrealtimebehaviour}. For all experiments we fix the sampling variance as $\Sigma = 1.0$. To prevent the control sequences from violating joint constraints during forward rollouts and execution on the robot, we clamp the control splines to only sample within a domain that satisfy joint constraints.

        \begin{table}
            \centering
            \begin{tabular}{>{\bfseries}l c c c c}
            \toprule
            % \rowcolor{gray!20}
            Experiment & $K$ [N/m] & $C$ [Ns/m] & $\mu$ & $R$ [Ns/m]\\
            \midrule
            Hefting & 500 & 0 & N/A & N/A\\
            Rubbing & 100 & 1 & \textbf{0.4} & 2.0\\
            Pinching & \textbf{800} & \textbf{10} & N/A & N/A\\
            Contouring & 500 & 0 & 0 & 0\\
            \bottomrule
            \end{tabular}
            \caption{\textbf{Contact Model Parameters. }Soft contact model parameters for all experiment scenarios. \textbf{Bold} parameters the unknown true parameters.}
            \label{tab:model_params}
        \end{table}
        % \vspace{-1em}
        
        \noindent\textbf{Mass Estimation Experiments. } The Franka end effector is free to move within 1 dimensional domain $\mathcal{W} = [0, 0.6] \ \mathrm{m}$ in the z direction with respect to world frame coordinates (at the Franka root), with bounded joint velocity constraints $v_{ee} \in [-1.0, 1.0] \ \mathrm{m/s}$. The robot starts at initial end effector position $p_{ee,0} = 0.2 \ \mathrm{m}$ with zero initial velocity. The Allegro hand attached to the Franka estimates the mass of a grasped tennis ball of mass $0.05$ kg (the results in this paper scale this value by a factor of 100). We initialize the variance of the mass parameter as $\Sigma_\theta = 10 \ \mathrm{kg}^2$. The contact model parameters in this scenario is shown in Table \ref{tab:model_params}.

        \noindent\textbf{Friction Estimation Experiments. } The Franka end effector is free to move within 2 dimensional domain $\mathcal{W} = [0, 1.0] \times [0.5, 1.0] \ \mathrm{m}$ with respect to a local wall frame coordinate system at the base of the wall, with bounded joint velocity constraints $v_{ee} \in [-1.0, 1.0] \ \mathrm{m/s}$ in both x and y directions. The robot starts at initial end effector position $p_{ee,0} = [0.5, 0.8] \ \mathrm{m}$ in wall frame coordinates (oriented $0.8 \ \mathrm{m}$ in the y direction from the root of the Franka, with its x axis pointing away from the wall) with zero initial velocity. The robot estimates the coefficient of friction between its end effector and the wall, with true parameter $\mu = 0.4$. We initialize the variance of the mass parameter as $\Sigma_\theta = 1$.  The contact model parameters in this scenario is shown in Table \ref{tab:model_params}.

        \noindent\textbf{Material Property Estimation Experiments. } A robot gripper enters a pregrasped state, where the gripper surface is positioned $0.4$ cm away from the surface of the ball object with stiffness coefficient $K = 800 \ \textrm{N/m}$ N/m, and dampening coefficient $C=10 \ \textrm{Ns/m}$. The gripper is free to move within gripper velocity constraints and prismatic joint constraints as per an Aloha 2 system.  We initialize the variance of the mass parameter as $\Sigma_\theta = [100.0 \ (\mathrm{N/m})^2, 10.0 \ (\mathrm{Ns/m})^2]$.

        \noindent\textbf{Shape Estimation Experiments. } A Franka is tasked with estimating the length $l = 0.126 \ \textrm{m}$ and width $w=0.05 \ \textrm{m}$ block. on a table surface. The robot is constrained in the world frame z axis such that is motion is planar about the surface of the table. The robot is initialized at end effector position $p_{ee} = [0.5, -0.1, 0.08]$ meters with respect to the world frame, and is constrained to a planar domain $\mathcal{W} = [0.3, -0.15] \times [0.8, 0.15] \ \mathrm{m}$, with velocity constraints set to $v_{ee} \in [-0.5, 0.5] \ \mathrm{m/s}$ for all axes. We initialize the variance of the mass parameter as $\Sigma_\theta = 10 \ \mathrm{cm}^2$. The contact model parameters in this scenario is shown in Table \ref{tab:model_params}.

\end{document}